\documentclass[Afour,sageh,times]{sagej}

\usepackage{moreverb,url}
\usepackage{textcomp}
\usepackage{fixltx2e}
\usepackage{multirow}
\usepackage{multicol}
\usepackage{dblfloatfix}
\usepackage{graphics}
\usepackage{pifont}
\usepackage[colorlinks,bookmarksopen,bookmarksnumbered,citecolor=red,urlcolor=red]{hyperref}

\newcommand\BibTeX{{\rmfamily B\kern-.05em \textsc{i\kern-.025em b}\kern-.08em
T\kern-.1667em\lower.7ex\hbox{E}\kern-.125emX}}

\def\volumeyear{2016}

\begin{document}

\runninghead{Hadi Hosseinabadi and Salcudean}

\title{Force Sensing in Robot-assisted Keyhole Endoscopy: A Systematic Survey}


\author{Amir Hossein Hadi Hosseinabadi and Septimiu E. Salcudean\affilnum{1}}

\affiliation{\affilnum{1}Robotics and Controls Laboratory (RCL), Electrical and Computer Engineering Department, University of British Columbia, Vancouver, British Columbia (BC), Canada}

\corrauth{Amir Hossein Hadi Hosseinabadi,
Robotics and Controls Laboratory (RCL), Electrical and Computer Engineering Department, University of British Columbia, Vancouver, British Columbia (BC), Canada,\\
V6T 1Z4, Canada}
\email{ahhadi@ece.ubc.ca}

\begin{abstract}
Instrument-tissue interaction forces in Minimally Invasive Surgery (MIS) provide valuable information that can be used to provide haptic perception, monitor tissue trauma, develop training guidelines, and evaluate the skill level of novice and expert surgeons.Force and tactile sensing is lost in many Robot-Assisted Surgery (RAS) systems. Therefore, many researchers have focused on recovering this information through sensing systems and estimation algorithms.

This article provides a comprehensive systematic review of the current force sensing research aimed at RAS and, more generally, keyhole endoscopy, in which instruments enter the body through small incisions. Articles published between January 2011 and May 2020 are considered, following the Preferred Reporting Items for Systematic reviews and Meta-Analyses (PRISMA) guidelines. The literature search resulted in 110 papers on different force estimation algorithms and sensing technologies, sensor design specifications, and fabrication techniques.
\end{abstract}

\keywords{Force and Tactile Sensing, Medical Robots and Systems, Force Control, Haptics and Haptic Interfaces, Telerobotics}

\maketitle

\section{Introduction}
In Minimally Invasive Surgery (MIS), surgical access is provided through small incisions or natural orifices in the body. A surgical instrument is operated by the surgeon for tissue manipulation. Compared to open surgery, MIS provides less tissue trauma, postoperative pain, patient discomfort, wound complications and immunological response stress \cite{Wottawa2016}, lower risk of infection \cite{Soltani-Zarrin2018} and blood loss \cite{Dai2017}, shorter hospital stay \cite{Bandari2020a}, faster recovery \cite{Lee2015}, and improved cosmetics \cite{Aviles2016} all of which lead to improved therapeutic outcome and efficiency \cite{Otte2016} and lower morbidity and morality \cite{Aviles2017} making MIS cost-effective \cite{Faragasso2014}. Nonetheless, the ergonomically cumbersome posture increases surgeon fatigue. The limited instrument dexterity and visual perception of the scene \cite{Haghighipanah2017, Haouchine2018}, and the non-intuitive hand-eye coordination due to fulcrum motion reversal decrease accuracy and contribute to surgeon fatigue \cite{Hadi2019}. The high level of psychomotor skills needed increases the operation time and require a longer learning curve \cite{Shahzada2016}. The sense of touch is reduced by friction in the access port and instrument mechanism.

In Robotic MIS (RMIS), the surgical instrument is controlled by a robotic manipulator and operated by a remote surgeon. 
The robotic operation restores hand–eye coordination \cite{Aviles2015a} and innovations in tool design improve dexterity leading to improved ergonomics that reduce surgeon fatigue \cite{Stephens2019,Bandari2020a}. The enhanced 3D surgical vision, automatic movement transformations, fine motions, filtering of physiological hand tremor and motion scaling lead to improved surgery precision \cite{Sang2017}. However, the surgeon is isolated from the surgical site by robotic manipulators that do not provide the haptic perception \cite{Juo2020}. This deprives the surgeon of a rich source of information. Thus, many studies are targeted towards the reconstruction and evaluation of haptic feedback.

Haptics can be either tactile or kinesthetic \cite{Okamura2009}. The tactile perception is through the cutaneous receptors in the skin which can sense, for example, texture or temperature \cite{Mack2012}. The kinesthetic force feedback is perceived by mechanoreceptors in the muscle tendons to detect force, position and velocity information about objects \cite{Juo2020}. Traditionally surgeons use palpation to characterize tissue properties, detect nerves and arteries \cite{Bandari2017}, and identify abnormalities such as lumps and tumors \cite{Lv2020, Puangmali2012}. Moreover, the surgeons rely on the sense of touch to regulate the applied forces. Excessive forces can lead to tissue trauma, internal bleeding, and broken sutures. Insufficient forces however can lead to loose knots and poor sutures. \cite{Sang2017,Li2016}.

Direct Force Feedback (FF) and Sensory Substitution (SS) are the most common approaches of presenting surgeons with force information. While the direct method provides the most intuitive interaction \cite{Li2017-Y}, it is the most challenging one to implement, as it requires a method of force sensing and a safe and robust teleoperation interface for force reflection. A compromise between transparency and stability of different teleoperation frameworks is reported by \cite{Hashtrudi-Zaad2001}. In sensory substitution, visual, auditory, or vibro-tactile signals provide haptic perception to the surgeon. While safety can be easily guaranteed, this method can cause discomfort, distraction, and cognitive overload. In general, visual methods are shown to be the most effective feedback modality. \cite{Abdi2020} 

In summary, the introduction of haptic perception is proven to decrease operation time \cite{Abiri2017}, facilitate training, improve accuracy, and enhance patient safety for novice surgeons in complex tasks \cite{Juo2020}. More experienced surgeons learn to infer force information from visual cues such as the tissue and instrument deformations and the stretch in sutures \cite{Aviles2017}. Additionally, force information can be used to automate surgical robot tasks in dynamic and unstructured environments \cite{Kuang2020}, to identify tissues in real time, to create tissue-realistic models and simulators for training \cite{Stephens2019}, and to perform surgical skills assessment \cite{Soltani-Zarrin2018}. 
\color{black}
\subsection{Comparison to the Existing Reviews}
An extensive review of haptic perception and its efficacy in RMIS is presented by \cite{Amirabdollahian2018}. This review concluded that while there is a consensus on the need for haptic and tactile feedback, no commercial system is yet available that addresses this need. More recently, \cite{ElRassi2020} presented a brief overview of haptic feedback in teleoperated robotic surgery. \cite{Overtoom2019} and \cite{Rangarajan2020} surveyed virtual haptics in surgical simulation and training. 
The latter followed the Preferred Reporting Items for Systematic reviews and Meta-Analyses (PRISMA) guidelines to identify the relevant literature. The authors similarly affirm the efficacy of haptic feedback in surgical education. None of the publications above review the developments in the field of force sensing and estimation.

\cite{Abdi2020} reviewed research since 2000 on the efficacy of haptic feedback in teleoperated medical interventions. The authors present a concise overview of the force-sensing literature with 44 references cited over a wide range of medical applications. Although the review provides a general understanding of the challenges and complexities in instrument-tissue force measurement, it is not a comprehensive presentation of the prominent developments and the articles were subjectively selected with no evaluation criteria. Additionally, the records were only classified based on the sensing technology and the sensor location; However, the instrument's dexterity level, the sensing Degrees of Freedom (DoFs), and the performance measures were not compared. A comparison of its references with the records cited in our review shows an overlap of only 20 out of 110 papers. \cite{Bandari2020a} reviewed tactile sensing literature over the past twenty years. It also includes some literature on force-sensing in neurosurgery and microsurgery procedures. Although the authors presented a comprehensive review with 121 references, a comparison of the included articles with the records in this paper shows an overlap of only 8 out of 110 papers which are mostly on developments related to the gripping force sensing. 

This article is a systematic review based on the PRISMA guidelines that expands on the sensor design requirements and presents the most recent developments in force sensing and estimation in keyhole endoscopy. 
We discuss how research has evolved over the past decade and provide suggestions for future research directions. The closest publications to our review are the surveys by Puangmali et al. \cite{Puangmali2008} and Trejos et al. \cite{Trejos2010} which were published about a decade ago, and therefore there are no overlapping papers with those reviews.
\color{black}
\begin{figure}[h]
    \centering
    \includegraphics[width = \linewidth]{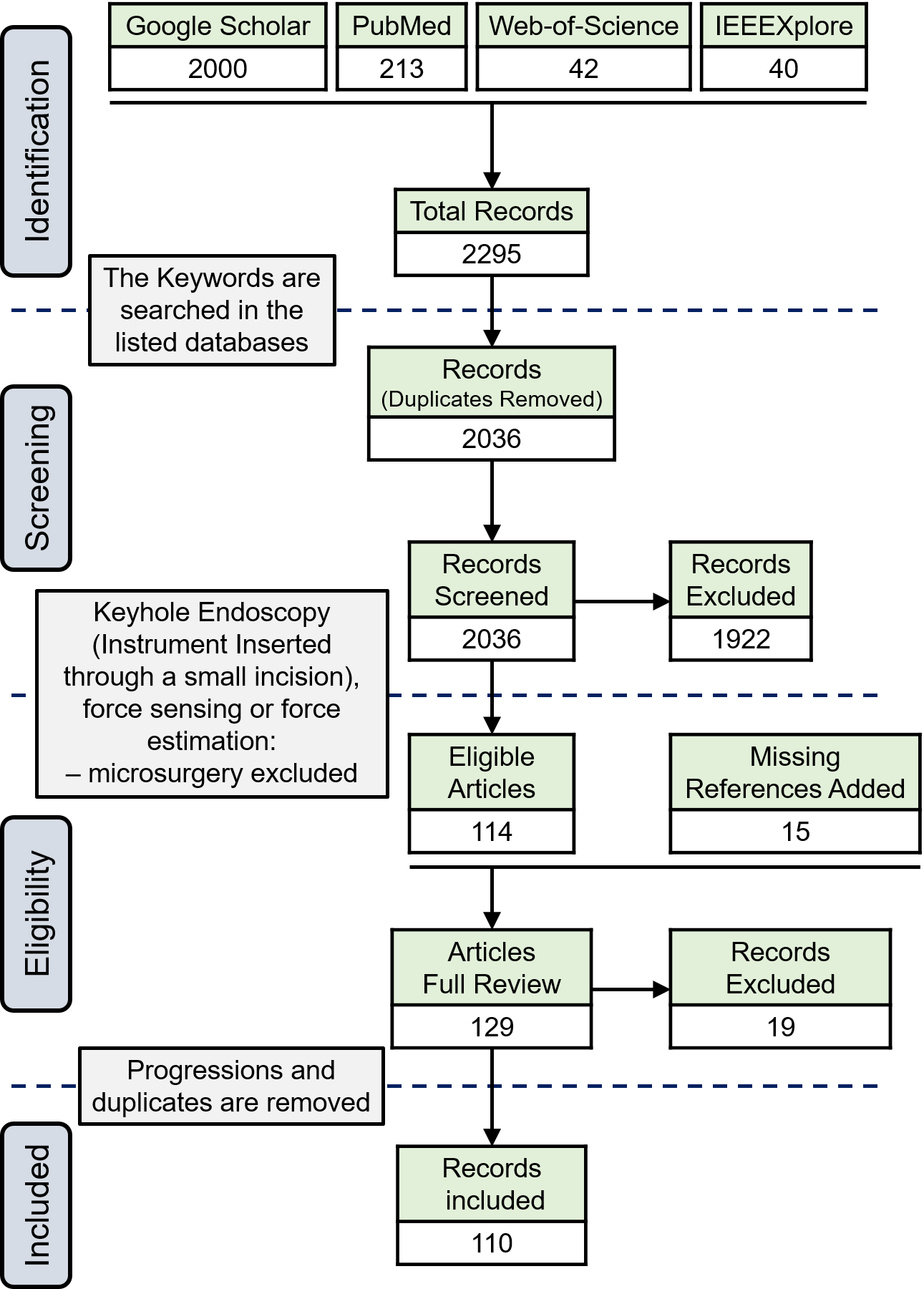}
    \caption{PRISMA flow diagram}
    \label{fig:PRISMA}
\end{figure}

\section{Methodology}\label{sec: Methodology}
A systematic survey was conducted by following the PRISMA guidelines (see Figure \ref{fig:PRISMA}) and it was based on Google Scholar, Web-of-Science, PubMed, and IEEE Xplore Digital Library repositories. The period for the review is over the past decade, from January 2011 until May 2020. The following keywords were used for identification: Force sensing, Kinesthetic, Tactile, Haptics, Minimally Invasive Surgery, MIS, MIRS, Robot-Assisted, RMIS, RAS, RAMIS, Laparoscopy, and Endoscopy. For every year, the first 20 pages of search results in Google Scholar were surveyed (total of 2000 records). The same approach was used for the identification of records through the other repositories (PubMed: 213, Web-of-Science: 42, and IEEEXplore: 40). For screening, the duplicates were removed and the identified records were skimmed through to mark the ones that are relevant to keyhole endoscopy. The articles that refer to force sensing in microsurgery, neurosurgery, and needle insertion were excluded because they involve a different set of requirements and challenges. Specifically, microsurgical instruments such as those used in neurosurgery and retinal surgery \cite{Gonenc2017} have a much smaller diameter (less than 2 mm) and do not require an articulated wrist, which complicates the actuation system and sensors' power and signal routing. \textcolor{black}{Moreover, \cite{Bandari2020a} briefly discussed the force-sensing literature in microsurgery, neurosurgery, and needle insertion.} 114 articles were found eligible for a complete review. Throughout the review, the references of the selected papers were surveyed and the relevant articles that were not initially identified were added, thus increasing the total number of eligible records to 129. The work progressions and duplicate publications were removed to lead to the 110 articles included in this survey.
\begin{figure}[hbt!]
    \centering
    \includegraphics[width = 0.9\linewidth]{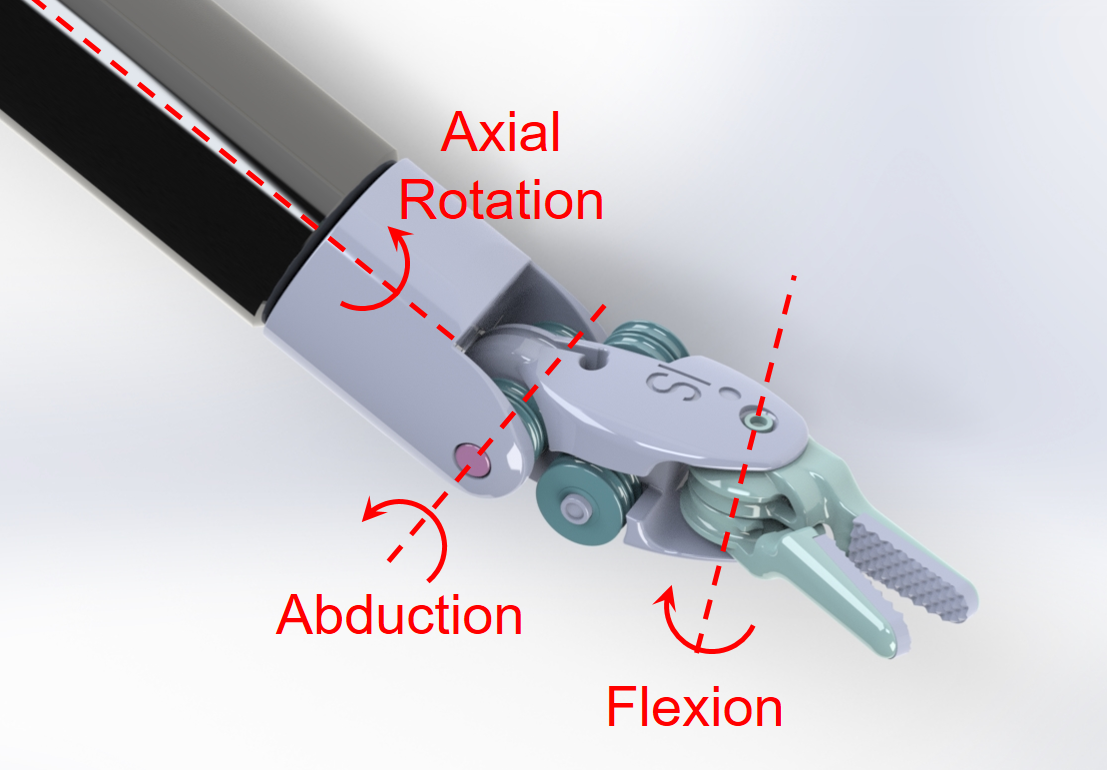}
    \caption{Instrument's degrees of freedom}
    \label{fig:actDoFs}
\end{figure}

\begin{figure}[]
    \centering
    \includegraphics[width = \linewidth]{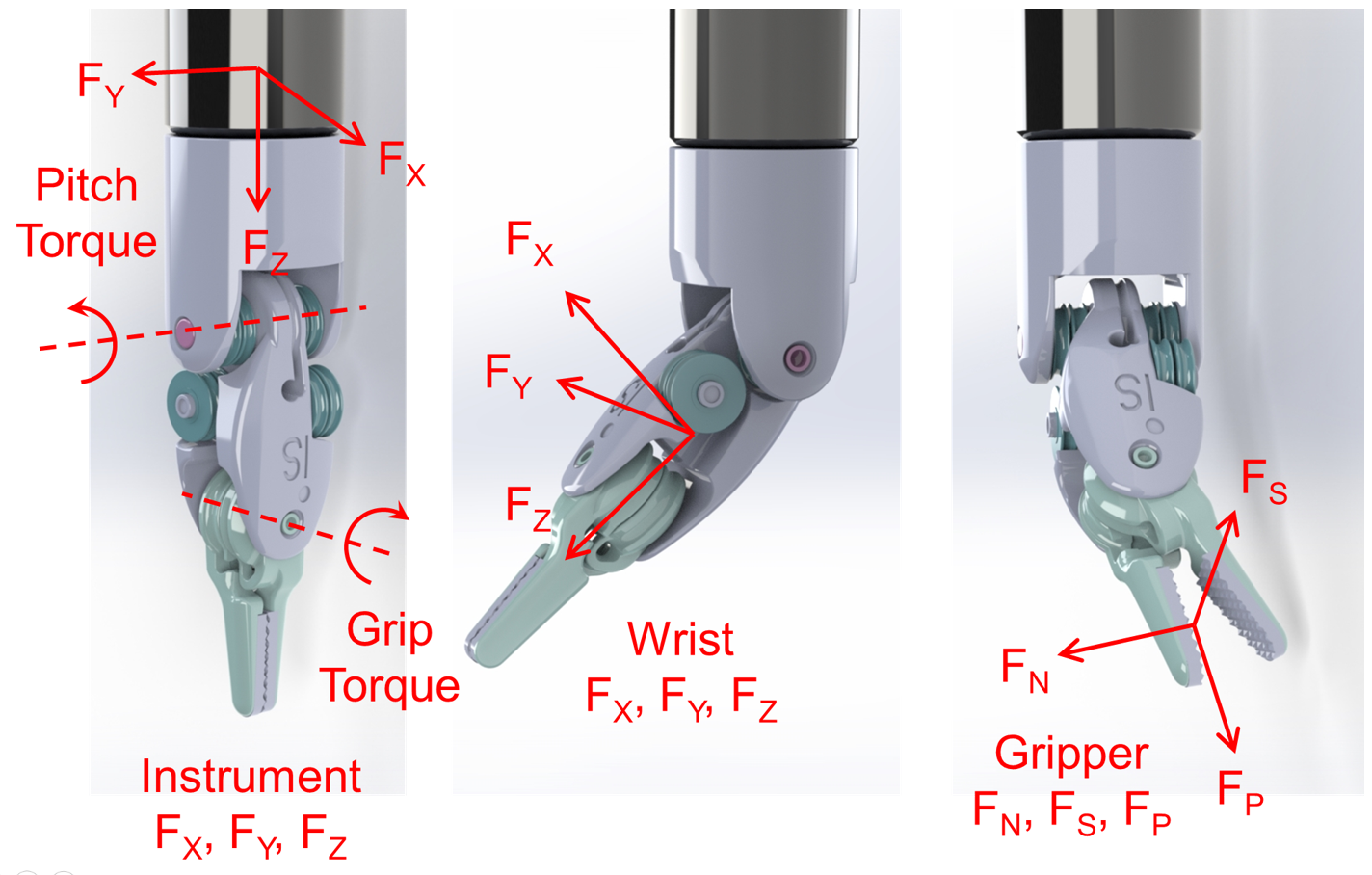}
    \caption{Sensing degrees of freedom depending on the sensor location}
    \label{fig:sensDofs}
\end{figure}

The included articles are tabulated for an easier comparison of the method, the sensor location, the sensing DoFs, the dexterity of the instrument under study, and the results. The Dexterity Index (DI) for different instruments is defined according to the Table \ref{tab:insDex} and Figure \ref{fig:actDoFs}. Depending on the sensor location, the sensing DoFs are defined as instrument or wrist tri-axial forces (F\textsubscript{X}, F\textsubscript{Y}, F\textsubscript{Z}) and moments (M\textsubscript{X}, M\textsubscript{Y}, F\textsubscript{Z}), and the gripper normal (F\textsubscript{N}), shear (F\textsubscript{S}), and pull (F\textsubscript{P}) forces as depicted in Figure \ref{fig:sensDofs}. In summarizing the results, the following acronyms were used: ACC: Accuracy, ERR: Maximum absolute error, MAE: Mean-Absolute-Error, NRMSE: Normalized-Root-Mean-Square-Error, RES: Resolution, RMSE: Root-Mean-Square-Error, RNG: Range, and SENS: Sensitivity. 

\begin{table}[hbt!]
    \small\sf\centering
    \caption{Dexterity index definition for MIS instruments \label{tab:insDex}}
    \begin{tabular}{p{2cm}p{4.75cm}}
    \toprule
    \centering Dexterity \newline Index (DI) & Instrument Functionality \newline and DoFs\\
    \midrule
    \centering- &   Standalone Testing \\
    \centering0 &   Palpation\\
    \centering1 &   Grasping\\
    \centering2 &   Grasping + Axial rotation \\
    \centering3	&   Grasping + Flexion \\
    \centering4	&   Grasping + Flexion \newline + Axial rotation \\
    \centering5	&   Grasping + Flexion + Abduction \\
    \centering6	&   Grasping + Flexion + Abduction \newline + Axial rotation \\
    \bottomrule
    \end{tabular}\\[10pt]
\end{table}

\section{Design Requirements}
\subsection{DoF, Range, Resolution, Accuracy, Bandwidth and Sampling Rate}
The grasping force, the instrument lateral and axial forces, and the axial torque are the most relevant DoFs to improve accuracy and provide an effective haptic experience in MIS applications \cite{Wee2016,Soltani-Zarrin2018,Bandari2020a}. Deformations in the sensor structure or displacements in its components are the physical surrogates that are monitored for force estimation. Thus, there are always trade-offs between the sensor’s structural rigidity, resolution and sensitivity, and range \cite{Puangmali2012}. The grip force can reach up to 20~N in da~Vinci instruments during needle handling or knot-tying \cite{ONeill2018, Bandari2020a}; however, pinch forces as large as 4~N can cause damage to delicate tissue \cite{Hong2012, Abiri2019a}. The maximum allowable suture pull force is 4-6~N \cite{Spiers2015, Dai2017}. The optimal kinesthetic force range suggested for MIS applications is $\pm$10~N in all directions and $\pm$20~N for grasping \cite{Khadem2016, Wee2017}. 
No requirement on bending moments and axial torque is specified in the literature \cite{Hadi2019}. Resolutions of 0.06~N \cite{Wang2014, Choi2017} and 0.2~N \cite{Soltani-Zarrin2018} are suggested for FF and SS schemes, respectively. The human just-noticeable difference (JND) is 10\% \cite{Wang2014, Kim2017} in the range of 0.5 to 200 N increasing to 15-27\% below 0.5 N \cite{Hwang2017} which can be considered as a requirement on the sensor accuracy. The human’s temporal resolution is 320~Hz for force discrimination and up to 700 Hz for vibration detection \cite{Puangmali2012}. However, the desired bandwidth of the force sensor is usually dictated by the application (FF, SS, vibration detection, etc.) and desired noise and resolution performance. A sample rate of 500~Hz is considered appropriate for direct force feedback applications \cite{Jones2017}. Sample rates as low as 30~Hz can be effective in visual SS modality.

\subsection{Size, Mass, and Packaging}
Surgical instruments are inserted into the body through a cylindrical port of 12-15 mm in diameter \cite{Spiers2015, Li2015a}. The outside diameter of the instrument is desired to be less than 10 mm \cite{Shi2019}. The sensor should be lightweight to not significantly increase the instrument inertia. The operation rooms are filled with equipment that can cause electromagnetic interference, and the electrocautery tools operate at high voltages \cite{Lim2014, Seok2019}. Thus, the sensors require insulation for electrostatic protection and shielding against electromagnetic interference \cite{Pena2018}. The sensors that enter the body also require sealing against humidity and debris ingression \cite{Trejos2014}.

\subsection{Sterilizability}
Surgical instruments are cleaned and sterilized for reuse; the former refers to removing debris from the device and the latter is the elimination of microorganisms that can cause disease \cite{Trejos2014}. The common sterilization methods are plasma and gamma radiation, the use of chemicals (alcohol, ethylene oxide or formaldehyde), and steam sterilizations \cite{Bandari2020b}. Steam sterilization is the fastest and the most preferred method \cite{Soltani-Zarrin2018} which is performed in an autoclave at 120-135\textdegree C, 207 kPa and 100\% humidity for 15-30 minutes \cite{Spiers2015,Zhao2015}. This harsh environment can be destructive to many transducers, signal conditioning electronics, wire insulations, bondings, and coatings. 

\subsection{Biocompatibility}
The sensors for use in MIS must abide by ISO 10993 which entails a series of standards for evaluating the biocompatibility of medical devices \cite{Trejos2014}. For biocompatibility electrical components often require coatings that interfere with sterilizability \cite{Spiers2015}. 

\subsection{Adaptability and Cost}
Instruments are disposed after 10 to 15 uses due to accelerated cable fatigue \cite{Anooshahpour2014, Kim2018a, Xue2018}. The EndoWrist instruments retail at \$2k-\$5k \cite{Spiers2015}. If the sensor is integrated into the instrument and is to be disposed, it should not increase the instrument price significantly. An adaptable solution that can be easily used on different instruments is desirable. 

\section{Location}
\begin{figure}[hbt!]
    \centering
    \includegraphics[width = 0.95\linewidth]{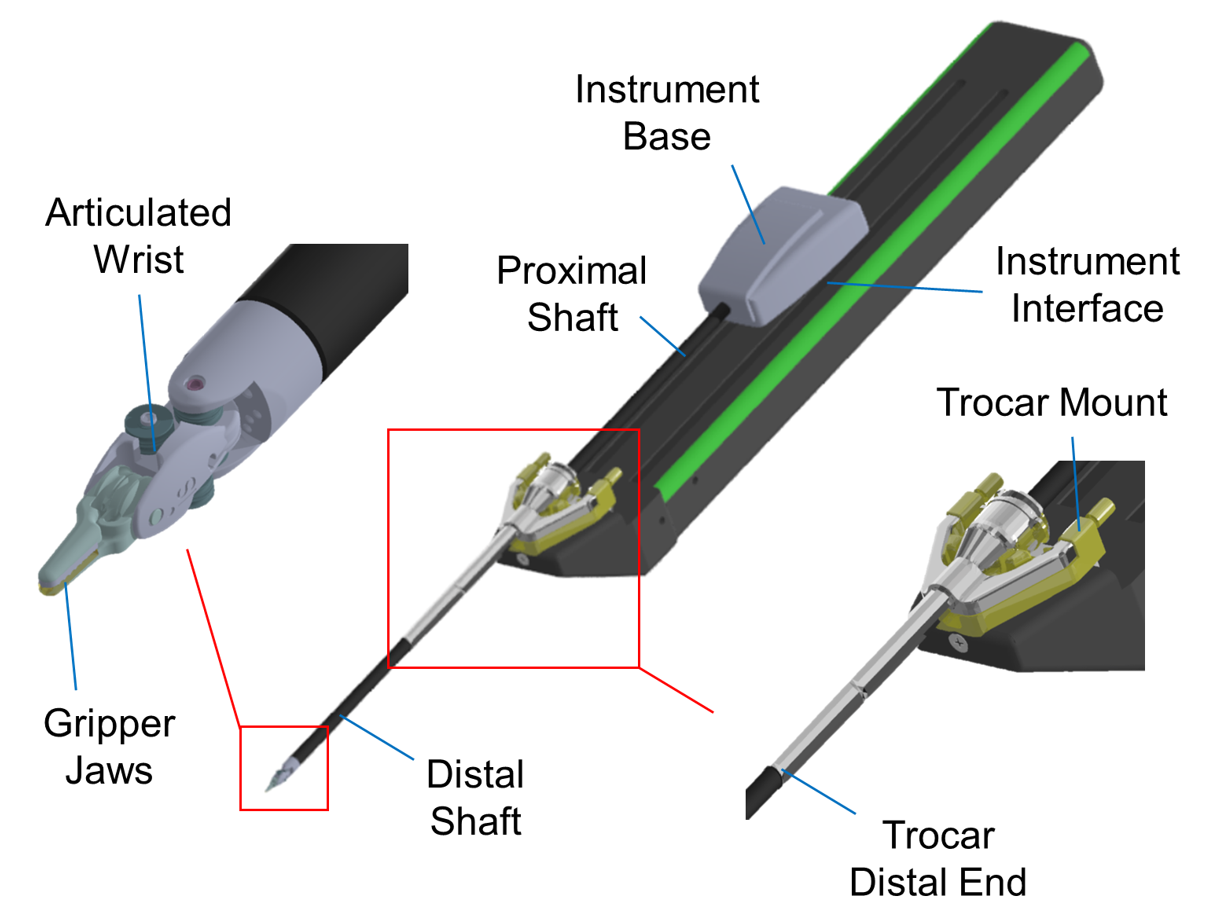}
    \caption{The options for sensor location}
    \label{fig:sensLoc}
\end{figure}

The sensors can be placed in the instrument mounting interface, the instrument base, proximal (outside the body) and distal (inside the body) shafts, the actuation mechanism (cables/rod), the trocar mount and its distal end, the articulated wrist, and the gripper jaws (Figure \ref{fig:sensLoc}). 
\begin{figure*}[t]
    \centering
    \includegraphics[width = \linewidth]{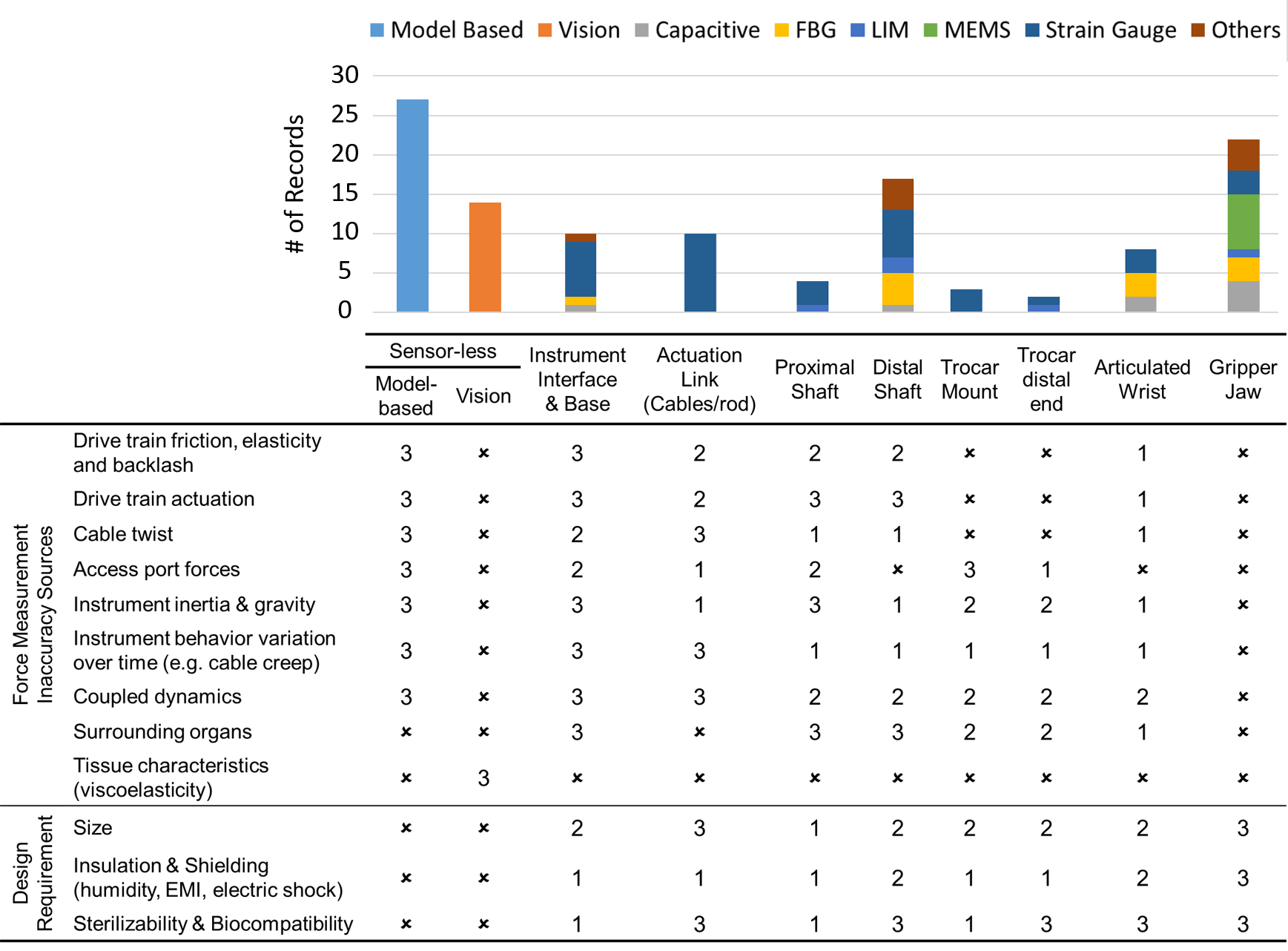}
    \caption{Overview}
    \label{fig:techSummary}
\end{figure*}
While the size, sterilizability, biocompatibility, and insulation requirements are more relaxed for the sensors placed outside the body, these locations are more prone to the factors causing sensor inaccuracy. The sensors at the instrument interface, base, and proximal shaft can have the electronics isolated from the patient \cite{VanDenDobbelsteen2012}. The sensors in the instrument shaft can gain high precision in the lateral direction, but experiments \cite{Maeda2016, Lv2020} showed that they do not provide high resolution in the axial direction unless the structure is modified to amplify axial strains. The sensors in the instrument shaft and trocar cannot independently measure the gripping force \cite{He2014} and measuring cable tensions cannot provide information on the axial force. The sensors integrated into the trocar and instrument interface are usually adaptable \cite{Wang2014}.

Sensors placed at the gripper jaw provide the most accurate readings and have the most stringent design constraints. They are difficult to fabricate, package, mount \cite{Kim2018b, He2014}, and shield \cite{Seok2019, Wang2014} and have limited adaptability which makes them cost-prohibitive for disposable instruments \cite{Xue2019}. Additionally, the electronics are usually placed away from the transducer which deteriorates the Signal to Noise Ratio (SNR) \cite{Suzuki2018}. Placing the force sensor at the grasper may also conflict with functional requirements for monopolar or bipolar cautery instruments \cite{He2014}.

Figure \ref{fig:techSummary} summarizes the severity level of different sources that contribute to the sensing inaccuracy as a function of the sensor location (scale of 1 to 3; 1 is minimum, 3 is maximum and \ding{56} is no effect). It also compares how stringent the listed design requirements are for each sensor location (scale of 1 to 3; 1 is the least, 3 is the most and \ding{56} refers to not a requirement). The distribution of the records included in this survey as a function of the sensor locations and the sensing technologies are shown in the same figure. It is evident that the sensorless techniques have the majority of publications over the past ten years. Additionally, the Micro-Electro-Mechanical (MEM) and Fiber Brag Grating (FBG) technologies have been widely adopted in the fabrication of miniature transducers that can be integrated into the gripper jaws. An overview of different transduction technologies is presented in the next section.
\begin{figure}[h]
    \centering
    \includegraphics[width = \linewidth]{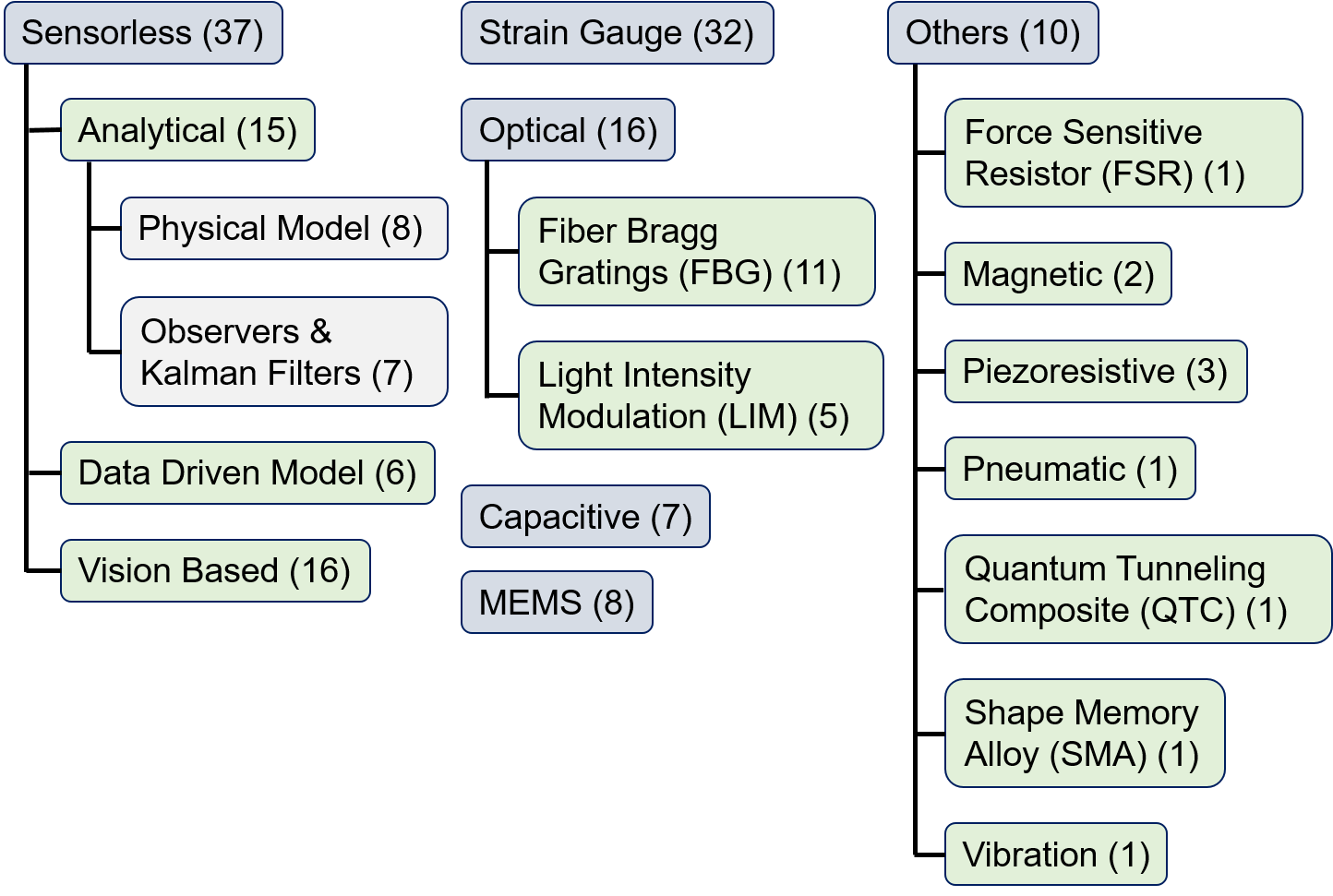}
    \caption{The sensing technologies}
    \label{fig:sensTech}
\end{figure}

\begin{table*}[t]
\small\sf\centering
\caption{Sensorless force estimation: model-based \label{Tab: ModelBased1}}
\begin{tabular}{p{1.8cm}p{7.6cm}p{1.5cm}p{1.6cm}p{2.8cm}}
\toprule
Author&Method&Sensing\newline DoFs&Instrument\newline/ DI&Results\\
\midrule
1- \cite{Li2013} & Pneumatic Actuation Muscles (PAM) were used in design of a custom forceps. Disturbance observers were used on the pneumatic actuation system and the robot joints. & Instrument tri-axial forces& Custom Developed (CD) RMIS \newline forceps / 5 & ERR $<$ 0.4N \newline RNG: 0-3.5N \\
\midrule
2- \cite{Tsukamoto2014} & Proposed a three step Robust Reaction Torque Observer (RRTO): 1) Cancel the error in estimated torque (overshoot correction) 2) Identify and compensate the inertial torque component in the drive train (inertia compensation) 3) Estimate the gripping torque. & Grip torque & CD RMIS \newline instrument \newline / 4
& Plot comparison, not \newline quantified.\\ 
\midrule
3- \cite{Anooshahpour2014} & Proposed two quasi-static models on cable dynamics (Pull \& Pull-Push) that take tendons friction \& compliance into account. A linear combination of the two models provided a close estimation of the output gripping torque. & Grip torque & EndoWrist Needle Driver / 6 & Plot comparison, not \newline quantified. \newline RNG: $\pm$ 40Nmm \\
\midrule
4- \cite{Lee2014} & The gripper was actuated by a pneumatic catheter balloon to provide a uniform gripping force. The pneumatic pressure was monitored to estimate the grip force. & Grip force & CD RMIS \newline instrument \newline / 5 & ERR $<$ 0.3N \newline RNG: 0-10N \\
\midrule
5- \cite{Zhao2015} & A wrist actuation design using planetary gears was proposed to decouple the motions in different DoFs. The motor currents were used to estimate the forces in static and dynamic scenarios. & Instrument lateral forces, \newline grip force & CD RMIS \newline instrument \newline / 5 & ERR $<$ 0.4N \newline
RNG: 0-2N \\
\midrule
6- \cite{Lee2015} & Proposed the compensation of the gripping torque which was experimentally identified as a function of the instrument posture. The experiments were on: EndoWrist 1) ProGrasp 2) Large Needle Driver 3) Dissecting forceps. & Grip force & EndoWrist instruments \newline / 6 & 1) ERR $<$ 10.69\% \newline 2) ERR $<$ 13.03\% \newline 3) ERR $<$ 16.25\% \\
\midrule
7- \cite{Haraguchi2015} & The articulated wrist was replaced by a machined spring. The instrument was pneumatically driven. A 3-DoF continuum model of the spring distal joint and the pneumatic pressure were used for force estimation. & Forceps \newline tri-axial & CD RMIS \newline instrument \newline / 6 & ERR $<$ 0.37N \newline
RNG: $\pm$ 5N \\
\midrule
8- \cite{Yoon2015} & Proposed the use of Sliding Perturbation Observer (SPO) to estimate the reaction force. The presented method compensated for the Coulomb friction. & \footnotesize{Grip force, \newline pitch torque, \newline instrument axial torque} & \footnotesize{EndoWrist ProGrasp \newline / 6} & \footnotesize{RNG: F\textsubscript{G}: 0-10N \newline Pitch torq.: $\pm$150Nmm \newline Axial torq.: $\pm$1Nm} \\
\midrule
9- \cite{Rahman2015} & Cascaded fuzzy logic in Sliding Mode Control with SPO (SMCSPO) to separate different types of disturbances. A rectified position information was defined and used to estimate the perturbation (grip force). & Grip force & EndoWrist ProGrasp \newline / 6 & RNG: 0-15N\\
\midrule
10- \cite{Li2016} & Proposed the use of an Unscented Kalman Filter (UKF) based on a dynamic model that considered cable properties, cable friction, cable-pulley friction, and a bounding filter. The proposed approach used motor currents and motor encoder readings. & Grip force & Raven-II 10 mm gripper \newline / 6 & ERR $<$ 50\% \newline RNG: 0-1N \\
\midrule
11- \cite{Anooshahpour2016} & Proposed the use of Preisach approach to model the input-output hysteretic behavior in a da Vinci instrument. & Grip force & EndoWrist Needle Driver / 6 & ERR $<$ 0.6N \newline RNG: 0-6.5N \\
\midrule
12- \cite{Sang2017} & Developed and identified a dynamic model for the Patient Side Manipulator (PSM) of the daVinci Standard system and the surgical instrument. The identified model was used for external force estimation. & Instrument tri-axial forces & EndoWrist Needle Driver / 6 & ERR $<$ 0.1N \newline RNG: $\pm$1.5N \\
\midrule
13- \cite{Haghighipanah2017} & Evaluated two approaches for force estimation on the 3\textsuperscript{rd} link of the Raven-II system: 1) Further expanded on the approach in \cite{Li2016} by adding cable tension estimation.  2) The force was estimated by measuring the cable stretch using a linear encoder. & Instrument axial force & Raven-II 10 mm gripper \newline / 6 & \footnotesize{ERR 1) $<$ 4N, 2) $<$ 3N \newline RNG: 0-10N \newline \#2 Provided better estimation at lower forces} \\
\midrule
14- \cite{Li2017-Y} & Used the Gaussian Progress Regression (GPR) supervised learning approach because of its ability to deal with uncertainties and nonlinearity. The model inputs were motors encoder, velocity, and current. & Grip force & Raven-II 10 mm gripper \newline / 6 & ERR $<$ 0.07N \newline RNG: 0-1N \\
\bottomrule
\end{tabular}\\[10pt]
\end{table*}

\begin{table*}[h]
\small\sf\centering
\caption{Sensorless force estimation: model-based - continue.\label{Tab: ModelBased2}}
\begin{tabular}{p{1.8cm}p{7.6cm}p{1.5cm}p{1.6cm}p{2.8cm}}
\toprule
Author&Method&Sensing\newline DoFs&Instrument\newline/ DI&Results\\
\midrule
15- \cite{Xin2017} & Developed the dynamic model of one jaw by using the Benson model to describe the dry friction. The parameters were experimentally identified for an instrument designed based on the concept in \cite{Zhao2015}. & Grip force & CD RMIS \newline instrument \newline / 5 & ERR $<$ 0.25N \newline RNG: 0-2.5N \\
\midrule
16- \cite{ONeill2018} & Evaluated motor current command and measurement, and differential gearbox as proximal torque surrogates and used Neural Networks (NNs) to estimate the distal gripping torque considering all three surrogates as inputs. & Grip force & daVinci Si \newline Maryland grasper \newline / 6 & ERR $<$ 0.37N \newline RNG: 0-11N \\
\midrule
17- \cite{huang2018} & Proposed the use of NNs optimized by a Genetic Algorithm (GA) for force estimation. The model inputs were the motors' positions, velocities, and currents. & Grip force & CD RMIS \newline instrument / 5 & ERR $<$ 0.06N \newline RNG: 0-1.6N \\
\midrule
18- \cite{Takeishi2019} & Suggested the use of pneumatic actuators and NN for force estimation. Low accuracy in abrupt forces was reported. All the analysis was model-based in MATLAB. & - & Simulation & RNG: 0-10 N\\
\midrule
19- \cite{Abeywardena2019} & A NN architecture with LSTM was proposed that used motors currents as the inputs. The model was trained for different stages of no grasp, closing, and opening. & Grip force & EndoWrist ProGrasp \newline / 6 & ERR $<$ 0.4N \newline RNG: 0-20N \\
\midrule
20- \cite{Stephens2019} & The performance of NNs, decision tree, random forest, and support vector machine models were compared in the angle and gripping torque estimation of each jaw. It concluded that the NN estimations were reliable when trained and tested on each jaw, on the same tool, and withing the frequency of the training data. & Grip force & EndoWrist ProGrasp \newline / 6 & ERR $<$ 0.07 N \newline RNG: 0-5.5N \\
\midrule
21- \cite{Wang2019a} & Proposed an external force estimation method based on cable-tension disturbance observer and the motion control strategy. & Grip force & CD RMIS \newline instrument \newline / 5 & ACC $>$ 85\% \newline RNG: 0.1-2N \\
\bottomrule
\end{tabular}\\[10pt]
\end{table*}

\section{Sensing Technologies}
Sensing technologies and the corresponding number of articles in this survey is shown in Figure \ref{fig:sensTech}.

\subsection{Sensorless}
\begin{table*}[h]
\small\sf\centering
\caption{Vision-based force estimation \label{Tab: VisionBased1}}
\begin{tabular}{p{1.8cm}p{7.6cm}p{1.5cm}p{1cm}p{3.3cm}}
\toprule
Author&Method / Task&Sensing\newline DoFs& Stereo \newline / Mono & Results\\
\midrule
1- \cite{Martell2011} & Image processing algorithms were utilized for suture strain estimation by identifying the suture line and tracking the displacement of markers. The achieved resolution in strain estimation was two orders of magnitudes smaller than the known strain to failure of most suture materials (20+\%). / Suture pull & Force magnitude & Mono & \footnotesize{Strain resolution of 0.2\% and 0.5\% was achieved in one-marker tracking on stationary suture and two-marker tracking on moving suture, respectively.}\\
\midrule
2- \cite{Kim2012} & The soft-tissue deformation was obtained by processing the stereoscopic depth image as a surface mesh. It was compared against the original organ shape from pre-operative images. A spring damper model was used for force estimation. / Tissue push  and pull & Gripper \newline tri-axial forces & Stereo & No results were presented \\
\midrule
3- \cite{Noohi2014} & A virtual template, based on assuming soft tissue local deformation to be a smooth function, was used to estimate the tissue deformation without a-priori knowledge of its original shape. The force magnitude was estimated by using a biomechanical model. / Tissue push & Gripper \newline tri-axial forces & Mono & In force magnitude: \newline ERR $<$ 0.12N \newline RMSE = 0.07N \newline RNG: 0-2.5N\\
\midrule
4- \cite{Faragasso2014} & A force sensing device composed of a linear retractable mechanism and a spherical visual feature was installed on the endoscope. The force was estimated as a function of the size of the spherical feature in the image. / Palpation & Instrument axial force & Mono & RES: 0.08N \newline RMSE = 0.13N \newline RNG: 0-1.96N \\
\midrule
5- \cite{Aviles2014} & The method used a 3D lattice to model the deformation of soft tissue. An RNN estimated the force by processing the information provided by the 3D lattice and the surgical tool motion. / Tissue push & Tissue \newline normal force & Stereo & MAE = 0.05N \newline RMSE = 0.062N \newline RNG: 0-3N \\
\midrule
6- \cite{Aviles2015a} & The RNN's full feedback architecture in \cite{Aviles2014} was replaced by local and global feedback. The RMSE and computation time were improved. / Tissue push & Tissue \newline normal force & Stereo & RMSE = 0.059N \newline RNG: 0-3N \\
\midrule
7- \cite{Aviles2015b} & The network in \cite{Aviles2015a} was upgraded to a recursive neural network LSTM based architecture which improves the force estimation accuracy. / Tissue push & Tissue normal force & Stereo & RMSE = 0.029N \newline RNG: 0-3N \\
\midrule
8- \cite{Otte2016} &  The tissue deformations from Optical Coherence Tomography (OCT) and the instrument trajectories were used as inputs to a Generalized Regression Neural Network (GRNN) to estimate the instrument-tissue forces. / Tissue push & Force magnitude & OCT Scanner & RMSE = 3mN \newline RNG: 0-20mN \\
\midrule
9- \cite{Aviles2016} & Evaluated the effect of dimensionality reduction on the performance of the RNN+LSTM architecture proposed in \cite{Aviles2015b}. It showed that implementation of a Probabilistic Principal Component Analysis (PPCA) significantly reduced dimension (75\% reduction) and improved accuracy. / Tissue push & Tissue normal force & Stereo & ERR $<$ 2\% \newline RMSE = 0.02N \newline RNG: 0-3N \\
\midrule
10- \cite{Giannarou2016} & The tissue deformations were estimated by finding stereo-correspondences based on tissues salient features and the use of probabilistic soft tissue tracking and thin-plate splines (TPS). The displacements were used to estimate forces based on a biomechanical model. / Tissue push & Force magnitude & Stereo &  MAE = 0.07N \newline RNG: 0-0.8N \\
\midrule
11- \cite{Aviles2017} & This was an extension to \cite{Aviles2016} where the proposed RNN+LSTM architecture was extended to three-axis force components. Z-axis was normal to the tissue, X and Y axes were planar with the tissue surface. / Tissue push & Tissue \newline tri-axial forces & Stereo & \footnotesize{RMSE: All DoF $<$ 0.02N \newline RNG: F\textsubscript{x}: $\pm$0.6N, F\textsubscript{y}: $\pm$2N, F\textsubscript{Z}: $\pm$6N} \\
\midrule
12- \cite{Hwang2017} & Same as \cite{Aviles2015b} with a deeper network, fully connected layers, and a sequence of mono 2D images as inputs. The results were on a sponge, a PET bottle, and a human arm with changes of light and pose. / Push & Tissue (object) normal force & Mono & \footnotesize{RMSE: Sponge:0.05N, PET bottle:0.17N, Arm:0.1N \newline RNG: Sponge:0-3N, PET bottle:0-7N, Arm:0-2N}\\
\midrule
13- \cite{Haouchine2018} & A biomechanical map of the organ shape was built on-the-fly from stereoscopic images. It used 3D reconstruction and meshing techniques.  / Tissue push  and pull & Force magnitude & Stereo & Plot comparison, \newline not quantified. \\
\bottomrule
\end{tabular}\\[10pt]
\end{table*}
\begin{table*}[h]
\small\sf\centering
\caption{Vision-based force estimation - continue \label{Tab: VisionBased2}}
\begin{tabular}{p{1.8cm}p{7.6cm}p{1.5cm}p{1cm}p{3.3cm}}
\toprule
Author&Method / Task&Sensing\newline DoFs& Stereo \newline / Mono & Results\\
\midrule
14- \cite{Gessert2018} & Took an undeformed reference volume and a deformed sample volume from OCT as inputs into a Siamese 3D-CNN architecture and output a 3D force vector. The results were compared with CNNs that take the difference or the sum of the undeformed and deformed volumes, and a CNN that takes 2D projected surface images as inputs. / Tissue push & Tissue \newline tri-axial forces & OCT Scanner & In force magnitude: \newline MAE: 7.7 mN \newline RMSE: 4.3 mN \newline RNG: 0-1 N\\
\midrule
15- \cite{Marban2018} & A semi-supervised learning model consisting of an encoder+LSTM network was suggested. The encoder learned a compact representation of the RGB frames from video sequences. The LSTM network used the tool trajectory information and the output of the encoder for force estimation. / Tissue push & Instrument tri-axial forces and moments & Mono &  RMSE: F\textsubscript{Z}:0.89 N \newline RNG: F\textsubscript{Z}:0-8 N\\
\midrule
16- \cite{Marban2019} & It used a CNN+LSTM architecture that processed the spatiotemporal information in video sequences and the temporal structure of tool data (the surgical tool-tip trajectory and its grasping status). / Tissue push  and pull & Instrument tri-axial forces and moments & Mono & \footnotesize{Plot comparison, not quantified. Concluded that both video sequences and tool data provide important cues for the force estimation.} \\
\bottomrule
\end{tabular}\\[10pt]
\end{table*}

\begin{table*}[b]
\small\sf\centering
\caption{Strain-gauge force sensing. \label{Tab: StrainGauge1}}
\begin{tabular}{p{1.8cm}p{5.5cm}p{2cm}p{2cm}p{4cm}}
\toprule
Author & Method / Location & Sensing\newline DoFs & Instrument \newline / DI & Results\\
\midrule
1- \cite{Jones2011} & Custom torque sensors were placed at the instrument interface between the driver and the driven knobs. / Instrument interface & Grip force, \newline Pitch torque, \newline Instrument axial torque & EndoWrist instruments / 6 & No results presented\\
\midrule
2- \cite{VanDenDobbelsteen2012} & A tension/compression load cell was installed in line with the actuating rod of the grasper. / Actuating rod & Grip force & Karl-Storz laparoscopic grasper / 1& ERR $<$ 0.09N \newline RNG: 0-2N \\
\midrule
3- \cite{Hong2012} & Custom grasper jaw with flexure hinges was designed to make a compliant structure. / Gripper & Gripper normal (F\textsubscript{N}) and pull (F\textsubscript{P}) forces & Standalone testing / 1
& RES: F\textsubscript{P}:43mN, F\textsubscript{N}:7.4mN \newline
RMSE: F\textsubscript{P}=95mN, F\textsubscript{N}=37mN \newline
RNG: $\pm$5N \\
\midrule
4- \cite{Baki2012} & Strain gauges were installed onto a custom-designed flexure out of Titanium fabricated by EDM. / Distal shaft & Instrument \newline tri-axial forces & Standalone testing / 0 & ERR $<$ 4\% \newline RES: 5mN \newline RNG: $\pm$ 2N \\ 
\midrule
5- \cite{He2014} & Custom designed sensors for measuring cable tension were installed at the instrument base. / Instrument base (cable tension) & Gripper normal and 3-DoF \newline forces& MicroHand robot instrument / 6 & ERR $<$ 0.4N \newline RNG: F\textsubscript{x},F\textsubscript{y}: $\pm$3.5, F\textsubscript{Z}: $\pm$2N, F\textsubscript{G}: 0-11N \newline \footnotesize (CS at an external sensor) \\
\bottomrule
\end{tabular}\\[10pt]
\end{table*}

\begin{table*}[h]
\small\sf\centering
\caption{Strain-gauge force sensing - continue. \label{Tab: StrainGauge2}}
\begin{tabular}{p{1.8cm}p{5.5cm}p{2cm}p{2cm}p{4cm}}
\toprule
Author & Method / Location & Sensing\newline DoFs & Instrument \newline / DI & Results\\
\midrule
6- \cite{MoradiDalvand2014} & Strain gauges were installed on the lead-screw actuation mechanism and a sleeve. / Actuating rod \& Distal end of  a sleeve
& Instrument lateral (F\textsubscript{L}) and Grip force (F\textsubscript{g}) & \footnotesize{CD RMIS \newline instrument for \newline 5mm fenestrated inserts} / 2 & MAE: F\textsubscript{L} $<$ 0.05N, Dir. $<$ 3\textdegree \newline RMSE: F\textsubscript{L}: $<$ 0.05N, Dir. $<$ 5.7\textdegree \newline RNG: F\textsubscript{L}: $\pm$ 1N, F\textsubscript{g}: 0-5N \\
\midrule
7- \cite{Wang2014} & An instrumented cover plate at the instrument interface and sensorized docking clamps at the trocar mount measured the z-axis and lateral forces, respectively. / Instrument interface \& trocar mount &  Instrument \newline tri-axial forces & EndoWrist instruments / 6 & RMSE $<$ 8\% \newline RNG: F\textsubscript{x},F\textsubscript{y}:$\pm$8N, F\textsubscript{Z}:$\pm$12N \\
\midrule
8- \cite{Talasaz2014} & Strain gauges were installed on the actuating cables and the RMIS instrument was attached to the robot flange through a 6 axis ATI Gamma F/T sensor. / Actuating Cables and instrument interface	& Instrument \newline tri-axial forces, axial and pinch torques, grip force	& EndoWrist Needle Driver \newline / 6 &	ERR: F\textsubscript{X},F\textsubscript{Y},F\textsubscript{Z} $<$ 0.12N \\
\midrule
9- \cite{Yu2014} & Small-size six-dimensional force/torque sensor with the structure of double cross beams. / Articulated wrist & Wrist 6DoF \newline forces \& \newline moments & CD RMIS \newline instrument / 5 & ERR $<$ 4.5\% \newline RNG: F\textsubscript{X},F\textsubscript{Y},F\textsubscript{Z}: 10N, \newline M\textsubscript{X},M\textsubscript{Y}: 150Nmm, M\textsubscript{Z}: 50Nmm \\
\midrule
10- \cite{Trejos2014} & Strain gauges were installed on the rod that actuated the grasper and on the distal end of the instrument shaft. / Actuation rod and distal shaft & Instrument lateral and \newline grasping forces &	Manual Laparoscopic grasper / 1 & ERR: F\textsubscript{X},F\textsubscript{Y},F\textsubscript{G} $<$ 0.2N \newline
RNG: F\textsubscript{X},F\textsubscript{Y}: $\pm$5N, F\textsubscript{G}: 0-17N \\
\midrule
11- \cite{Spiers2015} & Custom torque sensors were placed at the instrument interface between the driver and the driven knobs. / Instrument interface & Grip force, \newline Pitch torque, \newline Instrument axial torque, &	EndoWrist Needle 
Driver / 6	& RNG: $\pm$ 6N \\
\midrule
12- \cite{Li2015a}	& Strain gauges were installed on a custom-designed tripod flexure. / Distal shaft & Instrument \newline tri-axial & Standalone testing / 0	& ERR: F\textsubscript{X}, F\textsubscript{Y} $<$ 1\%, F\textsubscript{Z} $<$ 5\% \newline  
RNG: F\textsubscript{X}, F\textsubscript{Y}: $\pm$1.5N, F\textsubscript{Z}: $\pm$3N\\
\midrule
13- \cite{Li2015b} & Strain gauges were integrated into a custom-designed Flexural-hinged Stewart platform. / - &	6 DoF forces and torques & Standalone Testing / 0	& RES: F\textsubscript{X}, F\textsubscript{Y}: 0.08N, F\textsubscript{Z}: 0.25N
M\textsubscript{X},M\textsubscript{Y},M\textsubscript{Z}: 2.4Nmm \newline
RNG: F\textsubscript{X}, F\textsubscript{Y}, F\textsubscript{Z}: $\pm$30N, M\textsubscript{X},M\textsubscript{Y},M\textsubscript{Z}: $\pm$300Nmm \\
\midrule
14- \cite{Ranzani2015} & Two custom holders with integrated ATI F/T sensors were designed for the instrument and the fulcrum point. / Instrument interface and fulcrum point & Instrument \newline tri-axial forces &  MIS laparoscopic grasper / 1 & ERR $<$ 2.7\% \newline RNG: $\pm$4N \\
\midrule
15- \cite{Maeda2016} & An ATI Mini40 force sensor was mounted to the shaft of the instrument. / Proximal shaft &	Instrument \newline lateral and \newline axial forces, \newline axial torque & CD RMIS \newline laparoscopic forceps /6 & Sensor  performance  not \newline quantified. \\
\midrule
16- \cite{Khadem2016} & Integrated a tension/compression load cell inline with the lead-screw actuation and a 6-axis ATI Mini45 at the instrument base. / Actuating rod and instrument interface & Gripper pull force, \newline grip force & CD RMIS \newline laparoscopic grasper &  ERR: F\textsubscript{G} $<$ 0.5N \newline RNG:  F\textsubscript{G}: 0-5N \\
\midrule
17- \cite{Wee2016,Wee2017} & Presented a force-sensing sleeve with 4 strain gauges adaptable to standard MIS instruments. / Distal shaft & Instrument \newline tri-axial forces, axial torque & MIS Laparoscopic Grasper / 1 &	RES: 0.2N \newline RMSE: F\textsubscript{X}, F\textsubscript{Y} $<$ 0.088N \newline
RNG: F\textsubscript{X}, F\textsubscript{Y}:  $\pm$5N \\
\midrule
18- \cite{Barrie2016} & A tension/compression load cell was installed in line with the actuating rod of the grasper. / Actuating rod & Grip force & Johan fenestrated grasper / 1 & Sensor performance not \newline presented \\ 
\midrule
19- \cite{Seneci2017} & Proposed a disposable sensor clip for the gripper. The gripper was fabricated by Selective Laser Melting (SLM) and the sensor clip was 3D printed. /	Gripper &	Gripper normal force & Standalone testing / - & ERR $<$ 0.2N \newline
RNG: $\pm$5N \\
\midrule
20- \cite{Trejos2017} & Strain gauges were installed onto the proximal and the distal shafts. / Distal and proximal shafts & Instrument tri-axial forces &	MIS laparoscopic grasper / 1 &	ERR: F\textsubscript{X}, F\textsubscript{Y} $<$ 0.2N, F\textsubscript{z} $<$ 1.7N \newline
RNG: F\textsubscript{X}, F\textsubscript{Y}: $\pm$5N, F\textsubscript{Z}: $\pm$ 12N \\
\bottomrule
\end{tabular}\\[10pt]
\end{table*}

\begin{table*}[h]
\small\sf\centering
\caption{Strain-gauge force sensing - continue. \label{Tab: StrainGauge3}}
\begin{tabular}{p{1.8cm}p{5.5cm}p{2cm}p{2cm}p{4cm}}
\toprule
Author & Method / Location & Sensing\newline DoFs & Instrument \newline / DI & Results\\
\midrule
21- \cite{Li2017} & Extension on \cite{Li2015b} in which the sensor was integrated into the surgical instrument. / Articulated wrist	& Wrist 6DoF forces and moments & CD RMIS \newline instrument / 6 & RES: F\textsubscript{X}, F\textsubscript{Y}: 0.12N , F\textsubscript{Z}: 0.5N, M\textsubscript{X},M\textsubscript{Y},M\textsubscript{Z}: 7Nmm \newline
RNG: F\textsubscript{X}, F\textsubscript{Y}, F\textsubscript{Z}: $\pm$10N, M\textsubscript{X}, M\textsubscript{Y}, M\textsubscript{Z}: $\pm$160Nmm\\
\midrule
22- \cite{Kim2017}	& A 3 axis I-Beam force sensor using strain gauges were designed to replace the trocar support. / Trocar mount &	Instrument lateral, trocar axial friction &	EndoWrist instruments / 6 & \footnotesize{RMSE: F\textsubscript{X} $<$ 0.39N, F\textsubscript{Y} $<$ 0.20N, F\textsubscript{Z} $<$ 0.35N \newline RNG: F\textsubscript{X}, F\textsubscript{Y}: $\pm$15N, F\textsubscript{Z}: $\pm$10N}\\
\midrule
23- \cite{Schwalb2017} & It is similar to the overcoat method by \cite{Shimachi2003}. The instrument was mounted to an inner tube that was attached to a 6 axis F/T sensor. / Instrument interface & Instrument \newline tri-axial forces &	CD RMIS \newline instrument / 6 & RES: 0.09N \newline RNG: $\pm$9N \\
\midrule
24- \cite{Yu2018a}	& Axial load cells measured the cables tensions and a NN was used for friction compensation. / Actuating cable &	Gripper normal (F\textsubscript{N}) and shear (F\textsubscript{S}) forces & CD RMIS \newline instrument / 5 &	ERR: F\textsubscript{N} $<$ 10\%, F\textsubscript{S} $<$ 8\% \newline
RNG: F\textsubscript{N}: 0-2N, F\textsubscript{S}: $\pm$2.5N\\
\midrule
25- \cite{Kong2018} & Characterized the grip force over 50k grasps of one instrument using torque sensors at the instrument interface. Trained different NNs with an error threshold of 2 Nmm. The NN inputs were the proximal position, velocity, and torque measurements. / Instrument interface & Grip torque & EndoWrist Maryland grasper / 6	& ERR $<$ 2Nmm \\
\midrule
26- \cite{Karthikeyan2018}	& A custom flexure was designed and populated with strain gauges. / Articulated wrist & Wrist tri-axial forces & CD RMIS \newline instrument / 5 & RNG: 0-1.5N \\
\midrule
27- \cite{Novoseltseva2018} & The axial force was measured by a thin plate between the proximal shaft and the sterile adapter. The lateral forces were measured by a flexure at the trocar. / Proximal shaft and trocar distal end &	Instrument \newline tri-axial forces &	EndoWrist Needle Driver / 6	& \footnotesize{ERR: F\textsubscript{X} $<$ 0.4N, F\textsubscript{Y} $<$ 0.65N, F\textsubscript{Z} $<$ 0.63N \newline
RES: F\textsubscript{X}: 0.03N, F\textsubscript{Y}: 0.02N, F\textsubscript{Z}: 0.2N \newline RNG: F\textsubscript{X}, F\textsubscript{Y}: $\pm$19N, F\textsubscript{Z}: $\pm$12N} \\
\midrule
28- \cite{Pena2018} & Vapor-deposition fabrication techniques were used to directly print strain gauges on the instrument shaft. The material cost was \$0.09 per transducer. / Distal shaft & Instrument \newline lateral forces & EndoWrist Needle Driver \& Fenestrated grasper / 6 & ERR $<$ 0.8N \newline RNG: $\pm$5N \\
\midrule
29- \cite{Takizawa2018} & A disposable pneumatic cylinder with a strain gauge on its inner wall actuated the grasper. The transducer and the pneumatic pressure were used for force estimation. / Actuation system & Grip force & CD MIS \newline laparoscopic grasper & RNG: 0.1-0.25N\\
\midrule
30- \cite{Yu2018b}	& A custom gripper with double E-type beams flexure and populated with strain gauges was designed. / Gripper &	Gripper normal (F\textsubscript{N}), shear (F\textsubscript{S}), pull (F\textsubscript{P}) & Standalone testing / 1 &	RMSE: F\textsubscript{N}=23mN, F\textsubscript{S}=2.2mN, F\textsubscript{P}= 93mN \newline RES: 0.01N \newline RNG: $\pm$2.5N \\
\midrule
31- \cite{Wang2019b} & Combined the cable-drive dynamics, the cable tension measurement, and a Particle Swarm Optimization Back Propagation Neural Network (PSO-BPNN) to develop a joint torque disturbance observer. /	Cable tension &	Grip force & CD RMIS \newline grasper / 5 & ERR $<$ 0.25N \newline RNG: 0-2N \\
\midrule
32- \cite{Xue2019} & Four micro force sensors were used for cables tension measurement. The cable tension and a model of the cable drive system (with coupling and friction effects) were used to estimate the grasping forces. /	Cable tension &	Grip force & EndoWrist needle driver / 6 & ERR $<$ 0.4N \newline RNG: 0-12N \newline After stability is reached (Hysteresis effect) \\
\bottomrule
\end{tabular}\\[10pt]
\end{table*}

Sensorless refers to the case where the sensors used for force estimation are already inherent in the surgical robot \cite{Stephens2019}. In model-based approaches, the sensors are the encoders and the motor current measurements. In the vision-based techniques, the sensor is the visual feedback of the surgical site through mono or stereo cameras.

\subsubsection{Model-Based}
Model-based techniques can be categorized into 1) analytical models developed based on first principles, 2) disturbance observers and Kalman filters that utilize a dynamic model and the control loop commands and feedback signals, and 3) data-driven models which consider the instrument as a black-box and fit a quantitative model to a customized set of input and outputs. The model-based literature is summarized in Tables \ref{Tab: ModelBased1} and \ref{Tab: ModelBased2}. They include 15 analytical models, eight of which are physics-based and 7 studies use observers or Kalman filters. There are 5 articles on the use of data-driven models. 

The accurate dynamic model of the surgical instruments is challenging to obtain due to the many sources of nonlinearities e.g. friction, backlash \cite{Sang2017}, tendons compliance \cite{Anooshahpour2014} and creep \cite{Haghighipanah2017}, elastic deformations, actuators performance variations (the motors’ brush conductivity and change in the armature winding resistance) \cite{Li2016}, hysteresis \cite{Anooshahpour2016}, inertia, and gravity \cite{Wang2014}. Additionally, any model relies on a set of measurements (calibration or training set) that are usually taken at the beginning and used throughout the estimation. It is experimentally shown that the tool behavior changes with time which deteriorates the estimation accuracy \cite{Kong2018, Hadi2019}. The environmental parameters such as temperature and humidity can also affect the instrument characteristics \cite{Li2017-Y}. An alternative approach is the implementation of online adaptation and identification methods that are highly nonlinear, complex, and computationally demanding. This limits their effectiveness in real-time applications \cite{Anooshahpour2014, Li2016}. Dynamic modeling is particularly difficult in instruments with coupled degrees of freedom \cite{Zhao2015, Xin2017, ONeill2018}. Lee et al. \cite{Lee2015} showed that for the same input force by the surgeon, the grip force of the daVinci EndoWrist grasper can vary up to 3.4 times depending on its posture. As a result, despite the extensive research work, force estimations that rely on dynamic models do not provide highly reliable results yet, especially in the instrument’s lateral direction \cite{Fontanelli2017}. In comparison, the data-driven techniques based on supervised learning \cite{Li2017-Y, Stephens2019} provide more accurate force estimations.
\begin{table*}[t]
\small\sf\centering
\caption{Optical force sensing: LIM \label{Tab: OpticalLIM}}
\begin{tabular}{p{1.8cm}p{5.5cm}p{2cm}p{2cm}p{4cm}}
\toprule
Author&Method / Location&Sensing\newline DoFs& Instrument \newline / DI & Results\\
\midrule
1- \cite{Puangmali2012} & Presented a 3-axis force sensor with a flexible tripod structure, a stationary reflecting surface, and a pair of transmitting and receiving fibers per axis. The light source and photodetectors are remote or at the instrument base. / Distal shaft & Instrument \newline tri-axial forces &	Standalone testing / 0 & ERR: $<$ 5\%FS \newline
RES: 0.02N \newline
RNG: F\textsubscript{X}, F\textsubscript{Y}: $\pm$1.5N, F\textsubscript{Z}: $\pm$3N \\
\midrule
2- \cite{Ehrampoosh2013} &	Proposed an optical sensor design comprised of three Gradient-Index lenses (GRIN-lens) transmitting-receiving fiber-optic collimators, a flexible structure, and a reflective plate. / Distal shaft & Instrument \newline tri-axial forces	& Standalone testing / 0 & RNG: $\pm$6N.\\
\midrule
3- \cite{Fontanelli2017} & Used four optical proximity sensors to measure the deflection of the instrument shaft w.r.t the fixed trocar. The sensor was 3D-printed for proof of concept. / Trocar distal end & Instrument \newline lateral forces & Adaptable to any EndoWrist Instrument / 6	& ERR$<$12\% \newline RNG: $\pm$4N\\
\midrule
4- \cite{Hadi2019} & Optical force sensor comprising of an IR LED, a bicell photodiode, and a slit installed on the proximal shaft of the instrument. The proposed concept provided sub-nanometer resolution in deflection measurement. / Proximal shaft & Instrument \newline lateral forces & EndoWrist instrument / 6 & RMSE= 0.03N \newline RNG: $\pm$1N \\
\midrule
5- \cite{Bandari2020b} & A moving cylinder bends a fiber sitting on two fixed cylinders. Rate-dependent learning-based support-vector-regression was used for calibration. / Gripper & Grip force & CD MIS \newline laparoscopic grasper & ERR $<$ 0.2N \newline RES: 0.002N \newline
RNG: 0-2N \\
\bottomrule
\end{tabular}\\[10pt]
\end{table*}

\begin{table*}[t]
\small\sf\centering
\caption{Optical force sensing: FBG \label{Tab: OpticalFBG}}
\begin{tabular}{p{1.8cm}p{5.5cm}p{2cm}p{2cm}p{4cm}}
\toprule
Author&Method / Location&Sensing\newline DoFs& Instrument \newline / DI & Results\\
\midrule
1- \cite{Haslinger2013} & Similar to the DLR's miniature 6 axis force-torque sensor \cite{Seibold2005} with the strain gauges replaced by FBGs. The sensor structure was a Stewart platform to provide enhanced stiffness. /	Articulated wrist &	Wrist 6-DoF \newline forces and \newline moments & DLR MICA \newline instruments / 6 & ERR $<$ 18.6\% \newline RNG: F\textsubscript{X},F\textsubscript{Y},F\textsubscript{Z}: $\pm$6.9N,\newline M\textsubscript{X}, M\textsubscript{X}: $\pm$59.34Nmm, M\textsubscript{Z} = $\pm$49.53Nmm \\
\midrule
2- \cite{Lim2014} & Two optical FBGs were integrated into the forceps. Each fiber had two gratings for measuring the mechanical strain (on the surface) and for temperature compensation (at the center of the bending neutral axis).	/ Gripper &	Gripper normal force (F\textsubscript{N}) & CD MIS \newline laparoscopic grasper / 1 & RES: 1mN \newline RNG: 0-5N \\
\midrule
3- \cite{Song2014} &  3-axis force sensor with 4 longitudinal bendable beams populated with FBGs. Four other FBGs were integrated as references for temperature compensation. / Articulated wrist &	Wrist\newline tri-axial forces &	CD RMIS \newline instrument / 6 & ERR: F\textsubscript{X},F\textsubscript{Y} $<$ 0.1N, F\textsubscript{Z} $<$ 0.5N \newline RNG: $\pm$ 10N \\
\midrule
4- \cite{Yurkewich2014} &  Integrated 3 FBGs on the distal shaft and another FBG into the moving jaw of the grasper. /	Distal shaft and gripper & Instrument \newline lateral force (F\textsubscript{L}),  grip force (F\textsubscript{G}) & MIS arthroscopic \newline grasper / 1 & RMSE: F\textsubscript{L}= 0.213N, Dir = 4.37\textdegree, F\textsubscript{G} = 0.747N \newline
RNG: F\textsubscript{L}: $\pm$10N, F\textsubscript{g}: 0-20N \\
\midrule
5- \cite{Shahzada2016} & Four FBG sensors were attached to the instrument distal shaft in a two cross-section layout which is insensitive to the error caused by combined force and torque loads. / Distal shaft & Instrument \newline lateral forces & EndoWrist \newline Needle Driver / 6 & ERR $<$ $\pm$0.05N (95\% confidence interval) \newline
RES: 0.05N \newline
RNG: $\pm$2N \\
\midrule
6- \cite{Choi2017} & Custom flexure with three FBGs and an overload protection mechanism. The calibration algorithm was based on a two-layer NN. / Articulated wrist & Wrist \newline tri-axial forces & Standalone testing / - & ERR $<$ 0.06N \newline RNG: $\pm$12N \\
\midrule
7- \cite{Suzuki2018} & Four FBGs were integrated into the articulated wrist. The differential wavelength shift was used to achieve robustness to temperature and gripping force. / Articulated wrist & Wrist bending \newline forces and \newline moments & CD RMIS \newline instrument / 6 & RNG: F\textsubscript{X},F\textsubscript{Y}: $\pm$0.5N, \newline T\textsubscript{X},T\textsubscript{Y}: $\pm$50Nmm \\
\midrule
8- \cite{Soltani-Zarrin2018} & Two grasper designs with sliding stretchable T-shaped parts for enhanced axial strain. Axial FBGs were at the grasper’s bending neutral axes and its surface. / Gripper & Gripper normal (F\textsubscript{N}) and pull (F\textsubscript{P}) forces & Standalone
testing / - & ERR: \newline 
1: F\textsubscript{N} $<$ 0.57N, F\textsubscript{P} $<$ 0.78N \newline
2: F\textsubscript{N} $<$  0.81N, F\textsubscript{P} $<$ 0.9N \newline
RNG: F\textsubscript{N}: 0-10N, F\textsubscript{P}: 0-6N \\
\midrule
9- \cite{Xue2018} & The cable tensions were measured by FBGs pasted in the grooves on inclined cantilevers integrated into the Instrument base. /	Instrument base (Cable tension) & Grip force &	CD MIS \newline laparoscopic instrument with local \newline actuation / 5 &	ERR $<$ 0.5N \newline 
RES: 0.14N \newline
RNG: 0-15N \\
\midrule
10- \cite{Shi2019} & A force sensing flexure combining a Stewart base and a cantilever beam. The FBG was integrated along the central line of the flexure with its two ends fixed in grooves. / Distal shaft & Instrument 
axial force	& Standalone testing / - & No radial constraint: \newline
MAE: F\textsubscript{Z} $<$ 0.26N, RES: 21mN, RNG: F\textsubscript{Z}: 0-12N \newline
With radial constraint: \newline 
MAE: F\textsubscript{Z} $<$ 0.12N, RES: 9.3mN, RNG: F\textsubscript{Z}: 0-7N \\
\midrule
11- \cite{Lv2020} & The force sensor had a miniature flexure based on a Sarrus mechanism to achieve high axial sensitivity and a large measurement range. An FBG was tightly suspended along the central axis of the flexure. / Distal shaft & Instrument \newline axial force &	Standalone testing / - & ERR $<$ 0.06N \newline
RES: 2.55mN \newline
RNG: 0–5N \\
\bottomrule
\end{tabular}\\[10pt]
\end{table*}

\begin{table*}[t]
\small\sf\centering
\caption{Capacitive force sensing \label{Tab: Capacitive}}
\begin{tabular}{p{1.8cm}p{5.5cm}p{2cm}p{2cm}p{4cm}}
\toprule
Author&Method / Location&Sensing\newline DoFs& Instrument \newline / DI & Results\\
\midrule
1- \cite{Lee2014} &	Proposed a tendon drive pulley at the instrument base with an integrated torque sensor. A 3-axis force sensor was placed into the instrument shaft. Both sensors were based on the changes in the distance between the electrode and the ground. / Distal shaft and instrument base & Instrument \newline tri-axial forces, grip force & Custom Prototype of an MIS grasper / 1	& RNG: 0-0.5N \\
\midrule
2- \cite{Kim2015} & Two sensors consisting of a triangular prism shape and two capacitive-type transducers with an elastomeric polymer dielectric were integrated into the grasper. Molding was used to fabricate a prototype. / Gripper (tip) &	\footnotesize{Gripper normal (F\textsubscript{N}), shear (F\textsubscript{S}), pull (F\textsubscript{P}), grip (F\textsubscript{G}) forces} & CD RMIS \newline instrument for RAVEN-II / 6 &	
\footnotesize{
RES: F\textsubscript{P}=42mN, F\textsubscript{S}=72mN, F\textsubscript{N}=58mN, F\textsubscript{G}=46mN \newline
RMSE: F\textsubscript{P} $<$ 84mN, F\textsubscript{S} $<$ 0.114N, F\textsubscript{N} $<$ 73mN, F\textsubscript{G} $<$ 95mN \newline
RNG: F\textsubscript{P}: $\pm$2.5N, F\textsubscript{S} $\pm$2.5N, F\textsubscript{N} $\pm$5N, F\textsubscript{G}: 0-5N}
\\
\midrule
3- \cite{Kim2016}	& Two sensors with 3 electrodes and common grounds were integrated into the Gripper. The dielectric was air and the signal processing electronics were local. / Gripper (base) & \footnotesize{Gripper normal (F\textsubscript{N}), shear (F\textsubscript{S}), pull (F\textsubscript{P}), grip (F\textsubscript{G}) forces} & Custom prototype of an MIS Grasper / 1 & ERR: F\textsubscript{P} $<$ 0.42N, F\textsubscript{S} $<$ 0.15N, F\textsubscript{N} $<$ 0.92N \newline RNG: 0-8N \\
\midrule
4- \cite{Lee2016} & An extension on \cite{Lee2014} with the 3-axis force sensor moved into the articulated wrist and two capacitive torque sensors in the tendon drive pulleys of the gripper jaws. /	Articulated wrist and instrument base &	Wrist \newline tri-axial forces, grip Force &	CD RMIS \newline instrument
for RAVEN-II / 6 & NRMSE: F\textsubscript{X}=0.039 , F\textsubscript{Y}=0.056 , F\textsubscript{Z}=0.026 \newline
RNG: F\textsubscript{X}: $\pm$1N, F\textsubscript{Y}: $\pm$1N, F\textsubscript{Z}: $\pm$1.6N \\
\midrule
5- \cite{Kim2018a} & Proposed a capacitance sensing PCB with 8 electrodes and a CDC chip and a conductive deformable structure as the common ground.  A spherical cap was added to the sensor for testing it in a palpation task. / Articulated Wrist & 6-DoF forces and moments & CD RMIS \newline instrument for S-surge robot / 6 & \footnotesize{MAE: F\textsubscript{X}, F\textsubscript{Y}, F\textsubscript{Z}: 5.5\%FSO, M\textsubscript{X}, M\textsubscript{Y}, M\textsubscript{Z}: 2.7\%FSO \newline
RES:  F\textsubscript{X}: 0.22mN, F\textsubscript{Y}: 0.31mN, F\textsubscript{Z}: 0.11mN, M\textsubscript{X}: 0.47mNmm, M\textsubscript{Y}: 0.41mNmm, M\textsubscript{Z}: 0.17mNmm \newline
RNG: F\textsubscript{X}, F\textsubscript{Y}, F\textsubscript{Z}: 1N, M\textsubscript{X}, M\textsubscript{Y}: $\pm$20Nmm, M\textsubscript{Z}: $\pm$10Nmm}\\
\midrule
6- \cite{Kim2018b} & Extension of \cite{Kim2016} with a slight modification in the proximal gripper jaw and calibration scheme so that the combination of the capacitive transducers also resolved the axial torque about the gripper. / Gripper (base) & \footnotesize{Gripper normal (F\textsubscript{N}), shear (F\textsubscript{S}), pull (F\textsubscript{P}), grip (F\textsubscript{G}) forces, axial  torque (T\textsubscript{P}) }&	CD RMIS \newline instrument for 
S-surge robot / 6 &	\footnotesize{Integrated sensor: \newline
RES: F\textsubscript{N}=1.8mN, F\textsubscript{S}=2.0mN, F\textsubscript{P}=3.8mN \newline
MAE: F\textsubscript{N}:6.4\%, F\textsubscript{S}:3.4\%, F\textsubscript{P}:8.6\%, T\textsubscript{P}: 5.7\% \newline
RNG: F\textsubscript{P}, F\textsubscript{N}, F\textsubscript{S}:$\pm$5N, T\textsubscript{P}:$\pm$3Nmm}\\
\midrule
7- \cite{Seok2019} & Extension of \cite{Kim2018b} with humidity compensation. An AC shield was added to minimize the temperature effects on parasitic capacitance. An anodizing process was applied for electric insulation. The sensors range was extended to 20 N.	/ Gripper (base) & \footnotesize{Gripper normal (F\textsubscript{N}), shear (F\textsubscript{S}), pull (F\textsubscript{P}), grip (F\textsubscript{G}) forces, axial  torque (T\textsubscript{P}) } & CD RMIS \newline instrument for S-surge robot / 6 & 
RNG: $\pm$20 N \\
\bottomrule
\end{tabular}\\[10pt]
\end{table*}

\begin{table*}[t]
\small\sf\centering
\caption{MEMS force sensing \label{Tab: MEMS}}
\begin{tabular}{p{1.8cm}p{5.5cm}p{2cm}p{2cm}p{4cm}}
\toprule
Author&Method / Location &Sensing\newline DoFs& Instrument \newline / DI & Results\\
\midrule
1- \cite{Lee2013} & A thin-film capacitive sensor was fabricated using MEMS silk-screening technique on a PET film. /	Gripper	& \footnotesize{Gripper normal (F\textsubscript{N}), shear (F\textsubscript{S}), pull (F\textsubscript{P}) forces} & Standalone
testing / - & SENS: F\textsubscript{N}= 6.1\% , F\textsubscript{S}=10.3\%, F\textsubscript{P}= 10.1\% \newline
RNG: 0-12N \\
\midrule
2- \cite{Gafford2013} & The Pop-Up Book MEMS method was used to fabricate a grasper with a custom thin-foil strain gauge in a single manufacturing step. / Gripper	& Gripper normal force & Standalone
testing / - & RES: 30mN \newline RNG: 0-1.5N \\
\midrule
3- \cite{Kuwana2013} & A piezo-resistive sensor chip was manufactured by burying a substrate of several bent beams in different directions in resin. 	/ Gripper & \footnotesize{Gripper normal (F\textsubscript{N}), shear (F\textsubscript{S}), pull (F\textsubscript{P}), and grip (F\textsubscript{G}) forces} & MIS laparosc-opic grasper (Covidien; ENDOLUNG) / 1 &	No results presented \\
\midrule 
4- \cite{Gafford2014} & Used Printed-Circuit MEMS (PCMEMS) technique to develop a monolithic, fully-integrated tri-axial sensor with printed strain gauges. / - & Instrument \newline
tri-axial forces & Standalone testing / - & RES $<$ 2mN \newline RNG: F\textsubscript{X}, F\textsubscript{Y}: $\pm$500mN, F\textsubscript{Z}:$\pm$2.5N\\
\midrule
5- \cite{Nakai2017}& A 6-axis force-torque sensor chip composed of 16 piezo-resistive beams was fabricated by using ion beam etching and surface doping. The sensor is 2x2 mm and installed onto the grasper. /	Gripper & Gripper 6DoF forces and moments &	MIS
laparoscopic grasper / 1 & \footnotesize{RNG:\newline F\textsubscript{N}: 0-40N, F\textsubscript{S}: $\pm$12.5N, F\textsubscript{P}: $\pm$12.5N \newline
M\textsubscript{N}: $\pm$15Nmm, M\textsubscript{S}: $\pm$100Nmm, M\textsubscript{P}: $\pm$100Nmm}\\
\midrule
6- \cite{Dai2017} & Proposed a 3-axis capacitive force sensor with differential electrodes. The compressive load reduced the dielectric thickness, and shear forces changed the overlap area. The sensor was fabricated using MEMS lithography. / Gripper &	\footnotesize{Gripper normal (F\textsubscript{N}), shear (F\textsubscript{S}), pull (F\textsubscript{P}) forces }& EndoWrist
ProGrasp / 6 & RES: F\textsubscript{N}=55mN, F\textsubscript{S}=1.45N, F\textsubscript{P}=0.25N \newline
RNG: F\textsubscript{N}: 0-7N, F\textsubscript{S}:$\pm$11N, F\textsubscript{P}: $\pm$2N\\
\midrule
7- \cite{Rado2018} & Used deep reactive ion etching (DRIE) to fabricate a monolithic silicon-based 3-axis force piezoresistive sensor. The sensor was covered with a semi-sphere PDMS polymer. / Gripper &	\footnotesize{Gripper normal (F\textsubscript{N}), shear (F\textsubscript{S}), and pull (F\textsubscript{P}) forces, palpation} &	MIS laparoscopic grasper for Robin Heart robot / 1 & ERR $<$ 10\% \newline
RNG: 0-4N \\
\midrule
8- \cite{Tahir2018} & Presented a piezoelectric sensor fabricated using reduced Graphene oxide (rGO)-filled PDMS elastomer composite to measure the dynamic force. / Gripper & Gripper normal & MIS laparoscopic grasper / 1 &	RNG: 0.5-20N \\
\bottomrule
\end{tabular}\\[10pt]
\end{table*}

\begin{table*}[t]
\small\sf\centering
\caption{Other force sensing technologies\label{Tab: OtherTech}}
\begin{tabular}{p{1.8cm}p{5.5cm}p{2cm}p{2cm}p{4cm}}
\toprule
Author&Method / Technology / Location &Sensing\newline DoFs& Instrument \newline / DI & Results\\
\midrule
1- \cite{Vakili2011} & A Tekscan FlexiForce piezoresistive pressure sensor was integrated into one of the grasper jaws. /	Piezoresistive / Gripper & Gripper normal force & CD MIS \newline laparoscopic grasper / 1 & RNG: 0-4.4N\\
\midrule
2- \cite{Mack2012} & QTC Pills were integrated into a custom-designed support structure. / QTC / Instrument base & Instrument \newline tri-axial forces, grip force, \newline axial torque & CD RMIS \newline instrument / 6	& No results presented \\
\midrule
3- \cite{McKinley2015} & Palpation probe that could be added onto the instruments. It measured the axial compression of the sliding tip using a Hall Effect sensor. / Magnetic–Hall Effect Sensor / Distal shaft & Instrument axial force & EndoWrist instruments / 6 &	RES: 4mN \newline RNG: 0-1.6N \\
\midrule
4- \cite{Srivastava2016} & Superelastic Nitinol wires were used, instead of strain gauges, in two cross-sections arrangements for strain measurement. / SMA / Distal shaft & Instrument \newline lateral forces & EndoWrist Needle Driver / 6 &	RES: 55mN \newline
RMSE $<$ 32 mN \newline RNG: $\pm$4N \\
\midrule
5- \cite{Li2017-Lu} & Proposed a compact 3-axis force sensor design with integrated signal conditioning, power regulation, and ADC. The sensor used an array of force-sensitive resistors (FSR) with a mechanically pre-loaded structure. / FSR / Distal shaft &	Instrument \newline tri-axial forces &	Standalone testing / - & RES: 0.1N \newline
RNG: $\pm$8N\\
\midrule
6- \cite{Jones2017} & A 3D-printed grasper face with an embedded neodymium permanent magnet was attached to a soft silicone base that was mounted on top of a 3-axis hall effect sensor. Genetic Programming algorithm was used for sensor calibration. /	Magnetic–Hall Effect Sensor / Gripper & \footnotesize{Gripper normal (F\textsubscript{N}), shear (F\textsubscript{S}), and pull (F\textsubscript{P}) forces} & Standalone testing / - & Hysteresis ERR: \newline F\textsubscript{N} $<$ 1.58N, F\textsubscript{S}, F\textsubscript{P} $<$ 0.31N \newline
RNG: F\textsubscript{N}: 0-35N, F\textsubscript{S}, F\textsubscript{P}: $\pm$ 7N\\
\midrule
7- \cite{Bandari2017} & Proposed a hybrid sensor that used a piezoresistive transducer to measure normal force and LIM in optical fibers to estimate the tissue deformation. The sensor was out of silicon for biocompatibility. / Piezoresistive+LIM / Gripper &	Gripper normal force & Standalone testing / 1 & RNG: 0-2.5N \\
\midrule
8- \cite{Gaudeni2018} & Proposed the placement of a pneumatic balloon in a cavity on the surgical instrument or endoscopic camera. When needed, the membrane is inflated to contact the tissue. The pneumatic pressure and volume are monitored to estimate the force. / Pneumatic / Distal shaft & Palpation force & Standalone testing / - & ERR $<$ 0.24 N \newline RMSE: 0.11 N\newline RNG: 0-1.7 N\\
\midrule
9- \cite{Abdi2020} & Tekscan FlexiForce and A101 piezoresistive sensors were installed onto  the forceps via a custom 3D-printed mounting component. / Piezoresistive	 / Gripper & \footnotesize{Gripper normal (F\textsubscript{N}), shear (F\textsubscript{S}), pull (F\textsubscript{P}) forces}  & EndoWrist Cadiere \newline forceps / 6 & RNG: F\textsubscript{N}: 0-15N, F\textsubscript{S}: $\pm$44N, F\textsubscript{P}: 0-44N \\
\midrule
10- \cite{Kuang2020} & A slender shaft was excited by using a vibration motor. The structure’s tri-axial acceleration signals in time-domain showed discernible ellipse-shaped profiles when a force was applied. The acceleration profiles were characterized via regression to estimate the direction and magnitude of the applied force. /	Vibration monitor /	Proximal \& distal shaft & Instrument \newline tri-axial forces & Standalone testing / - & MAE: F\textsubscript{L} = 18\%, F\textsubscript{Z} = 6\% \newline
RES: F\textsubscript{L} = 0.098N, Dir= 10\textdegree.\newline
RNG: F\textsubscript{L}: 0-0.98N, F\textsubscript{Z}: 0-0.95N \\
\bottomrule
\end{tabular}\\[10pt]
\end{table*}
\subsubsection{Vision-Based}
The existing literature affirms that experienced surgeons use visual cues (tissue and instrument deformations and the stretch in the suture) as sensory feedback surrogates \cite{Martell2011, Noohi2014}. With the 3D stereoscopic view in robotic surgery providing depth information, and the developments in the available computational power (high-performance Graphic Processing Units-GPUs, cluster computers, and cloud platforms), a noticeable shift towards adoption of vision-based techniques was observed. While mechanical models of the tissue are presented, they are mostly complex and computationally expensive \cite{Aviles2016}. Most of the literature (Tables \ref{Tab: VisionBased1} and \ref{Tab: VisionBased2}) implement supervised learning architectures (Recurrent Neural Network (RNN) and Long-Short-Term-Memory (LSTM) \cite{Aviles2016, Marban2019}) with the video stream as inputs to estimate the instrument-tissue interaction forces. The vision-based techniques are robust to many sources of inaccuracy listed in Figure \ref{fig:techSummary}. However, they can be affected by the instrument occlusion, smoke and changes in the tissue properties, lighting conditions and camera orientation. The estimation update rate cannot be faster than the video frame rate which is usually 30 Hz. This limitation makes the vision-based approached not suitable for FF applications in which the control loop is desired to execute faster than 500 Hz \cite{Jones2017}. The current literature highlights that force estimation through video processing is easier in pushing tasks (characterized by smooth deformations) than those produced by pulling tasks that are characterized by irregular tissue deformations due to grasping \cite{Marban2019}.

\textcolor{black}{Force estimation based on using Optical Coherence Tomography (OCT) as the reference sensor is proposed by \cite{Otte2016} and \cite{Gessert2018}. OCT images provide volumetric data with a resolution of a few micrometers in which the tissue compression and subsurface deformations can be reflected. Thus, they contain a richer signal space compared to the mono and stereo visions that provide only the surface information.}

\subsection{Strain Gauge}
Strain gauges are the most commonly used transducers for force sensing \cite{Yu2018a}. They are accurate and small and can be designed in different configurations for multi-axis force sensing. Although the transducers are low cost with a price of \$10-\$25 per unit \cite{Srivastava2016}, they require special surface preparation, adhesives, and coatings for optimal performance that increases the assembly and integration cost \cite{Li2017-Lu}. When used in Wheatstone bridge arrangements, they require multiple wires for connection that makes packaging difficult for quick and seamless integration with surgical instruments \cite{Shi2019, Lv2020}. Strain gauges are highly influenced by electromagnetic noise and are not suitable for use close to other tools with strong magnetic fields (e.g. electrocautery) \cite{Choi2017, Xue2018}. They have low sensitivity and often require custom flexures or modifications in the load-carrying structure to amplify local strains \cite{Hadi2019}. Strain gauges are fragile and require mechanical overload protection \cite{Kuang2020}. They typically do not survive multiple sterilization cycles \cite{Shi2019} and lose repeatability. Trejos et al. \cite{Trejos2014, Trejos2017} conducted an extensive study on biocompatible adhesives and coatings that can withstand the harsh environment during steam sterilization. However, none of the combinations showed reliable measurements after seven cycles. Tables \ref{Tab: StrainGauge1} to \ref{Tab: StrainGauge3} summarizes the articles which utilize strain gauges or commercial strain-gauge based force sensors for MIS force sensing.

\subsection{Optical}
Optical methods use light intensity (e.g. photodiodes, phototransistors), frequency (e.g. Fiber Brag Gratings), or phase (e.g. interferometry) modulation for force measurement. The optical signal can be locally converted to electric signals, or be transferred with fibers for distal processing. Placing the electronics away from the instrument tip makes sterilization easier. The optical fibers are flexible, scalable, biocompatible, electrically passive, insensitive to electromagnetic noise and thus MRI compatible \cite{Pena2018}, durable against high radiation \cite{Lim2014}, immune to water \cite{Song2014}, corrosion-resistant \cite{Shi2019}, and low cost \cite{Bandari2020b}. However, optical fibers cannot be routed into small bending radii \cite{Trejos2014}. Additionally, The presence of small and intricate parts can make fabrication and assembly of fiber-based sensors costly \cite{Li2017-Lu}.

The Light Intensity Modulation (LIM) based sensors are vulnerable to light intensity variations due to the temperature or fiber bending \cite{Lv2020}. This can be improved by normalizing the optical signal against the emitted power \cite{Hadi2019}. Alternatively, a redundant strain-free fiber can be used to compensate for the effect of temperature or other sources of uncertainty \cite{Puangmali2012, Song2014}. Table \ref{Tab: OpticalLIM} summarizes the articles that address MIS force estimation based on LIM.

The FBG sensors are wave-length coded and insensitive to the changes in the light intensity. FBGs are very sensitive, have calibration consistency, and exhibit high SNR which provide repeatable and high-resolution strain measurements \cite{Shahzada2016}. Multiple gratings can be accommodated into one fiber \cite{Soltani-Zarrin2018} simplifying the design and signals routing. Thus, they are also used in shape sensing \cite{Lv2020}. Nonetheless, FBGs require interrogators for signal processing which the commercial systems cost between \$10k to \$100k \cite{Yurkewich2014}. The articles which used FBGs for MIS force estimation are summarized in Table \ref{Tab: OpticalFBG}.

\subsection{Capacitive}
Capacitive methods are attractive solutions for high resolution and compact force sensor designs. Compared to strain gauges, they provide limited hysteresis in microscale and increased sensitivity \cite{Sang2017}. However, they have a limited range \cite{Li2017-Lu} and are prone to thermal and humidity drift \cite{Bandari2020b}. The change in capacitance can be due to the change of the overlapping area or the distance between the two electrodes; the latter provides higher sensitivity and a more linear response  \cite{Kim2017}. The commercially available Capacitance to Digital Converter (CDC) chips such as the AD7147 from Analog Devices significantly simplify the signal processing, which was believed to be challenging for capacitive transducers \cite{Trejos2014}. However, they provide a low sampling rate. Table \ref{Tab: Capacitive} lists the articles that are based on the capacitive transduction principle. The sterilizability and biocompatibility of the existing literature are not evaluated.

\subsection{MEMS}
MEM sensors (Table \ref{Tab: MEMS}) operate based on the same physical principles discussed so far. However, MEM fabrication techniques such as deposition, etching, and lithography allow for the cost-effective production of small, fully-integrated, monolithic sensors \cite{Rado2018} with reduced lead time in prototypes and high throughput batch volumes \cite{Gafford2014}. Typically, MEMS do not require manual assembly, bonding, and alignment, and provide functional devices after the fabrication process \cite{Gafford2013}. By utilizing MEMS technology, it is possible to develop smart parts (e.g. grippers) with integrated sensing capability for micromanipulation \cite{Pandya2014}. Biocompatible coatings can be added to MEM sensors for biomedical applications.

\subsection{Other Technologies}
Piezoelectric transducers do not require an external power supply and have high stiffness \cite{Li2017-Lu}. However, they are subject to charge leakage and are not suitable for low frequency and static loads \cite{Sang2017}. They are also sensitive to temperature. Piezoresistive transducers used in force-sensitive resistors are scalable with low hysteresis and noise \cite{Bandari2017}. Nonetheless, their linear response is limited to a small range and they drift under constant load \cite{Juo2020}. They do not have the challenges associated with the integration of strain gauges, are relatively insensitive to humidity, and can be used in high temperatures above 170\textdegree C \cite{Li2017-Lu}. Shape Memory Alloys (SMAs) like Nitinol have a higher gauge factor compared to common metallic strain gauges and provide a larger range due to their stretchability. SMAs require an insulating coating for use on conductive surfaces. They can be clamped at two ends and do not require a backing material with special surface preparation. They are low cost and available at diameters as small as few microns \cite{Srivastava2016}. Quantum tunneling composite (QTC) pills are flexible polymers that act as insulators in resting-state but increase conductivity when compressed. They are very sensitive, provide a wide dynamic range, and are low cost (less than 1\$/pill). However, they are temperature sensitive and inaccurate in dynamic loading applications \cite{Novoseltseva2018}. Recently, vibration frequency and phase shifting due to an applied force have been measured for force estimation by the use of accelerometers. This approach is slow as it needs a few vibration cycles to generate stable and repeatable signals \cite{Kuang2020}. 

\section{Discussion and Conclusion}
In keyhole endoscopy, the surgeon's interaction with the surgical site is via slender instruments that are inserted into the body through small incisions. Despite the many benefits to the patient, the operation is more challenging for the surgeon due to the instruments' limited dexterity, fulcrum motion reversal, uncomfortable posture, and limited visual presentation. Additionally, the surgeon's force perception is affected by the forces between the instrument and the skin and the instrument's dynamics. The adoption of robotic and computer vision technologies resolves the limitations above and significantly improves the accuracy and efficiency in RMIS. However, most telesurgical systems completely isolate the surgeon from the tissue through the local/remote architecture of robotic telemanipulation. This deprives the surgeon of the rich information in palpation and direct interaction with the tissue. 
Without force feedback, the interaction of the surgeons with
the environment is not as intuitive as direct manipulation and therefore extensive training is required. 
Moreover, the lack of haptic feedback leads to a higher risk of errors and longer task completion time, up to 2 orders of magnitude in complex tasks \cite{Hannaford1991}, which may lead to higher surgery costs.

One active research stream in the field of robotic surgery is improving the sense of telepresence for the surgeons, also known as ``transparency". Direct force feedback is the most intuitive approach to improve transparency. For a fully transparent haptic experience, reliable interaction force sensing at the surgeon console and the instrument-tissue interface is required. This is in addition to a safe bilateral teleoperation architecture, and a local manipulator that is capable of reflecting the force commands, known as a haptic display. The extensive literature on haptic control indicates a trade-off between transparency and stability \cite{Hashtrudi-Zaad2001}. Alternatively, sensory substitution was proposed instead of haptic feedback, in the form of visual, auditory, or vibrotactile cues of force information. Although the safety can be easily guaranteed in systems with SS, it is not intuitive and can cause cognitive overload for the surgeon. The SS methods can also be used in MIS systems because no robotic manipulator is required for force reflection. The efficacy of different haptic feedback modalities in improving the surgical training and surgeon performance metrics has been studied extensively \cite{Amirabdollahian2018, Rangarajan2020, Overtoom2019, Abdi2020}. It is shown that a transparent haptic experience and visual feedback of force information improve the performance metrics and shorten the training time for novice surgeons in complex tasks. Apart from haptic feedback, the instrument-tissue interaction forces can be used for tissue damage monitoring, surgical skills assessment, development of surgical training guidelines, and to automate tasks.

Extensive research has been conducted to estimate or sense the instrument-tissue interaction forces. The functional requirements depend on the application. While it is not necessary to estimate the tissue forces precisely to provide an appropriate haptic experience \cite{Jones2011},  the bandwidth and sampling rate are important requirements to ensure low latency and smooth interaction with the remote environment. The sampling rate and bandwidth are less critical in SS.

Sensorless approaches utilize the information that is already available in the robotic manipulator; the axes positions and velocities, motors currents, and visual display of the surgical theatre. With the exponential growth, over the past decade, in the available computational power to researchers, data-driven approaches based on supervised learning \cite{Li2017-Y, Aviles2017} have been widely adopted. 
Among them, neural networks have shown promising results when trained and used on one particular instrument. However, they require a long and computationally-expensive training phase that is yet clinically-prohibitive. The training is based on a set of measurements at the beginning of the surgery that is used afterward for force estimation throughout the entire surgery. 
Proposed approaches that have an instrument's operational parameters as inputs, do not consider the variations between instruments and the change of instruments behavior throughout its use \cite{Kong2018}. Considering how the research direction has evolved over the past decade, experimentation with different model architectures, development of efficient training, and identification methods that can be automatically performed at the system start-up \cite{Spiers2015}, improving the computation time and incorporation of online adaptation techniques are attractive research areas to be further investigated. Moreover, all the existing literature uses the information at the patient manipulator for force estimation, but the inclusion of the operating parameters at the surgeon console may also improve the quality of force estimation.

The sensor design is another avenue towards collecting force data at the instrument tip. The sensor can be located inside or outside the body. The closer the sensor is to the instrument tip the more accurate the measurements are. However, the size, biocompatibility, sterilizability, insulation, and sealing requirements are more stringent when such an approach is followed. Design proposals for sensor integration into the instrument tip have limited adaptability because the instruments for different types of surgery have different shapes at the tip (e.g. EndoWrist cautery forceps, graspers, dissectors, needle drivers, etc.). Therefore, a custom sensor needs to be designed for every instrument which increases the development, fabrication, and maintenance costs.

A variety of transduction principles, including resistive, capacitive, optical, piezoelectric, and magnetic have been used in the development of sensing solutions for minimally invasive procedures. While strain gauges are still the most commonly used transducers, the study by Trejos et al. \cite{Trejos2017} showed that biocompatible adhesives and coatings can only survive a maximum of 6 steam sterilization cycles. Considering that the instruments are typically used 10 times before disposed, this would lead to a 40\% increase in the cost of the instruments with integrated strain gauges. Additionally, the installation of strain gauges is labor-intensive that contributes to increased cost.

A comparison of the publications summarized in this article with the surveys by \cite{Puangmali2008} and \cite{Trejos2010} indicates a noticeable shift towards utilizing FBG and MEMS technologies for the development of gripper integrated miniature sensors (Figure \ref{fig:techSummary}). 
FBGs are compact, sterilizable, biocompatible, electrically passive, and immune to electromagnetic noise. They provide high sensitivity with sub-micron resolution and can have multiple gratings embedded in one fiber which simplifies fiver and therefore optical signal management. 
While the commercial interrogators are expensive, there are signal conditioning solutions proposed to decrease the electronics cost \cite{Yurkewich2014}. The developments in MEMS technology have overcome the barrier of scale and cost in the fabrication of delicate miniature sensors. Additionally, MEMS sensors typically do not need manual assembly and can be integrated into the desired application after production. 

Another observable trend is the utilization of data-driven regression approaches for sensor calibration. Models based on neural networks and other supervised learning methods such as Gaussian Process Regression have shown unprecedented performance in handling nonlinearities and uncertainties in sensor calibration. Compared to the surgical instruments, the transducers show a more consistent response and do not need regular calibration unless removed and reintegrated. Efficient calibration approaches that can be quickly and automatically performed without operator intervention (e.g based on payload estimation) would benefit the RMIS systems.

\begin{acks}
Amir  Hossein  Hadi  Hosseinabadi  would  like  to  NSERC for his Canada Graduate Scholarship-Doctoral scholarship. Professor Salcudean gratefully  acknowledges  infrastructure  support  from  CFI  and funding  support  from  NSERC  and  the  Charles  Laszlo  Chair in  Biomedical  Engineering.  
\end{acks}
\bibliographystyle{sageH}
\bibliography{main.bib}




\end{document}